\documentclass[runningheads]{llncs}

\usepackage{graphics}
\usepackage{algpseudocode}
\usepackage{xcolor}
\usepackage{color}
\usepackage{fontawesome}
\usepackage{colortbl}
\usepackage{algorithm, algcompatible, float}
\usepackage{dsfont}
\usepackage{stackengine}
\usepackage{times}
\usepackage{epsfig}
\usepackage{graphicx}
\usepackage{amsmath}
\usepackage{amssymb}
\usepackage{multirow}
\usepackage{mathtools,xparse}
\usepackage{subcaption}
\usepackage{stmaryrd}
\usepackage{placeins}
\usepackage{pifont}
\usepackage{tabularx}
\usepackage{mdframed}
\usepackage{nopageno}
\usepackage{makecell}
\usepackage{arydshln}
\usepackage{booktabs}
\usepackage{enumitem}
\usepackage[skip=1ex,font=small,labelsep=period]{caption}
\usepackage{tabu}
\usepackage{etoolbox}
\usepackage{verbatim}
\usepackage{siunitx}
\usepackage{xspace}

\usepackage[accsupp]{axessibility}  

\DeclarePairedDelimiter\abs{\lvert}{\rvert}%
\DeclarePairedDelimiter\norm{\lVert}{\rVert}%

\makeatletter
\let\oldabs\abs
\def\abs{\@ifstar{\oldabs}{\oldabs*}}
\let\oldnorm\norm
\def\norm{\@ifstar{\oldnorm}{\oldnorm*}}
\makeatother

\definecolor{mygray}{gray}{.95}
\definecolor{headergray}{gray}{.85}
\definecolor{Gray}{gray}{0.85}
\newcolumntype{g}{>{\columncolor{Gray}}c}

\makeatletter
\def\BState{\State\hskip-\ALG@thistlm}
\makeatother

\algnewcommand\algorithmicforeach{\textbf{for each}}
\algdef{S}[FOR]{ForEach}[1]{\algorithmicforeach\ #1\ \algorithmicdo}

\newcommand{\mat}[1]{\mathbf{{#1}}}

\newcommand*\ie{\emph{i.e.}\xspace}
\newcommand*\eg{\emph{e.g.}\xspace}
\newcommand*\matr{\mathbf}
\newcommand*\trans{\text{T}}

\newcommand*\Point{\mathbf{p}}

\algrenewcommand\algorithmicrequire{\textbf{Input:\phantom{ll}}}
\algrenewcommand\algorithmicensure{\textbf{Output:}}

\newcommand{\Hom}{\matr{H}}
\newcommand{\Aff}{\matr{A}}
\newcommand{\Fund}{\matr{F}}
\newcommand{\Ess}{\matr{E}}

\begin{document}
\pagestyle{headings}
\mainmatter
\def\ECCVSubNumber{3998}  

\title{Relative Pose from SIFT Features} 

\titlerunning{Relative Pose from SIFT Features}
%
\author{Daniel Barath\inst{1} \and
Zuzana Kukelova\inst{2}}
\authorrunning{Daniel Barath and Zuzana Kukelova}
%
\institute{ETH Zurich, Computer Vision and Geometry Group, Switzerland\\
\email{danielbela.barath@inf.ethz.ch}\\
\and
Visual Recognition Group, FEE, Czech Technical University in Prague\\
\email{kukelova@cmp.felk.cvut.cz}}
\maketitle

\begin{abstract}
This paper proposes the geometric relationship of epipolar geometry and orientation- and scale-covariant, e.g., SIFT, features. 
We derive a new linear constraint relating the unknown elements of the fundamental matrix and the orientation and scale.
This equation can be used together with the well-known epipolar constraint to, e.g., 
estimate the fundamental matrix from four SIFT correspondences,
essential matrix from three,
and to solve the semi-calibrated case from three correspondences. 
Requiring fewer correspondences than the well-known point-based approaches (e.g., 5PT, 6PT and 7PT solvers) for epipolar geometry estimation makes RANSAC-like randomized robust estimation significantly faster. 
The proposed constraint is tested on a number of problems in a synthetic environment and on publicly available real-world datasets on more than \num{80000} image pairs.
It is superior to the state-of-the-art in terms of processing time while often leading more accurate results.
\end{abstract}

\section{Introduction}

This paper addresses the problem of interpreting orientation- and scale-covariant features, \eg SIFT \cite{lowe1999object} or SURF \cite{bay2006surf}, w.r.t. the epipolar geometry characterized either by a fundamental or an essential matrix.
The derived relationship is then exploited to design minimal relative pose solvers that allow significantly faster robust estimation than by using the traditional point-based solvers. 

Nowadays, a number of algorithms exist for estimating or approximating geometric models, \eg, homographies, using fully affine-covariant features. 
A technique, proposed by \cite{PerdochMC06}, approximates the epipolar geometry from one or two affine correspondences by converting them to point pairs.   
Bentolila et al.~\cite{Bentolila2014} proposed a solution for estimating the fundamental matrix using three affine features. 
Raposo et al.~\cite{Raposo2016,raposo2016pi} and Barath et al.~\cite{barath2018efficient} showed that two correspondences are enough for estimating the relative pose when having calibrated cameras. 
Moreover, two correspondences are enough for solving the semi-calibrated case, \ie, when the objective is to find the essential matrix and a common unknown focal length \cite{barath2017focal}. 
Guan et al.~\cite{guan2021relative} proposed ways of estimating the generalized pose from affine correspondences. 
Also, homographies can be estimated from two affine correspondences as shown in the dissertation of Kevin Koser~\cite{koser2009geometric}, and, in the case of known epipolar geometry, from a single correspondence \cite{barath2017theory}. 
There is a one-to-one relationship between local affine transformations and surface normals \cite{koser2009geometric,barath2015optimal}. 
Pritts et al.~\cite{Pritts2017RadiallyDistortedCT,pritts2018rectification} showed that the lens distortion parameters can be retrieved using affine features. 
The ways of using such solvers in practice are discussed in~\cite{barath2020making}. 

Affine correspondences encode higher-order information about the underlying the scene geometry. 
This is why the previously mentioned methods solve geometric estimation problems (\eg, homographies and epipolar geometry) using only a few correspondences -- significantly fewer than what point-based methods need. 
However, requiring affine features implies their major drawback.
Detectors for obtaining accurate affine correspondences, for example, Affine-SIFT \cite{morel2009asift}, Hessian-Affine or Harris-Affine \cite{mikolajczyk2005comparison}, MODS \cite{mishkin2015mods}, HesAffNet \cite{mishkin2018repeatability}, are slow compared to other detectors. 
Therefore, they are not applicable in time-sensitive applications, where real-time performance is required. 

In this paper, the objective is to bridge this problem by exploiting partially affine co-variant features.
The typically used detectors (\eg, SIFT and ORB) obtain more information than simply the coordinates of the feature points, for example, the orientation and scale. 
Even though this information is actually available ``for free'', it is ignored in point-based solvers. 
We focus on exploiting this already available information without requiring additional computations to obtain, \eg, expensive affine features.


\begin{figure}[t]
\centering
\begin{subfigure}{0.278\textwidth}
\centering
\includegraphics[trim={0 0 80mm 0},clip,width=\textwidth]{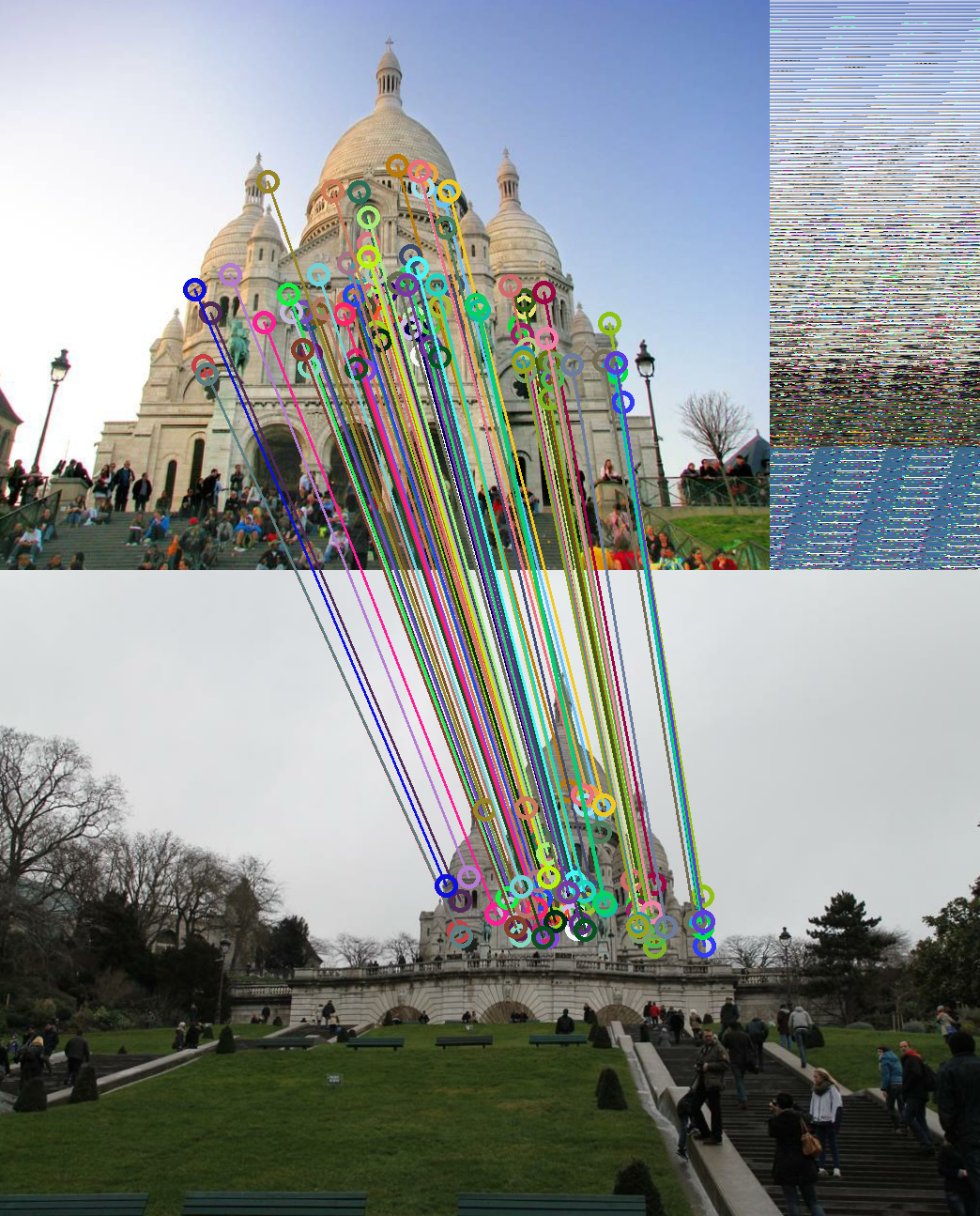}
\caption{Sacre Coeur}
\end{subfigure}
\begin{subfigure}{0.278\textwidth}
\centering
\includegraphics[trim={20mm 0 20mm 0},clip,width=\textwidth]{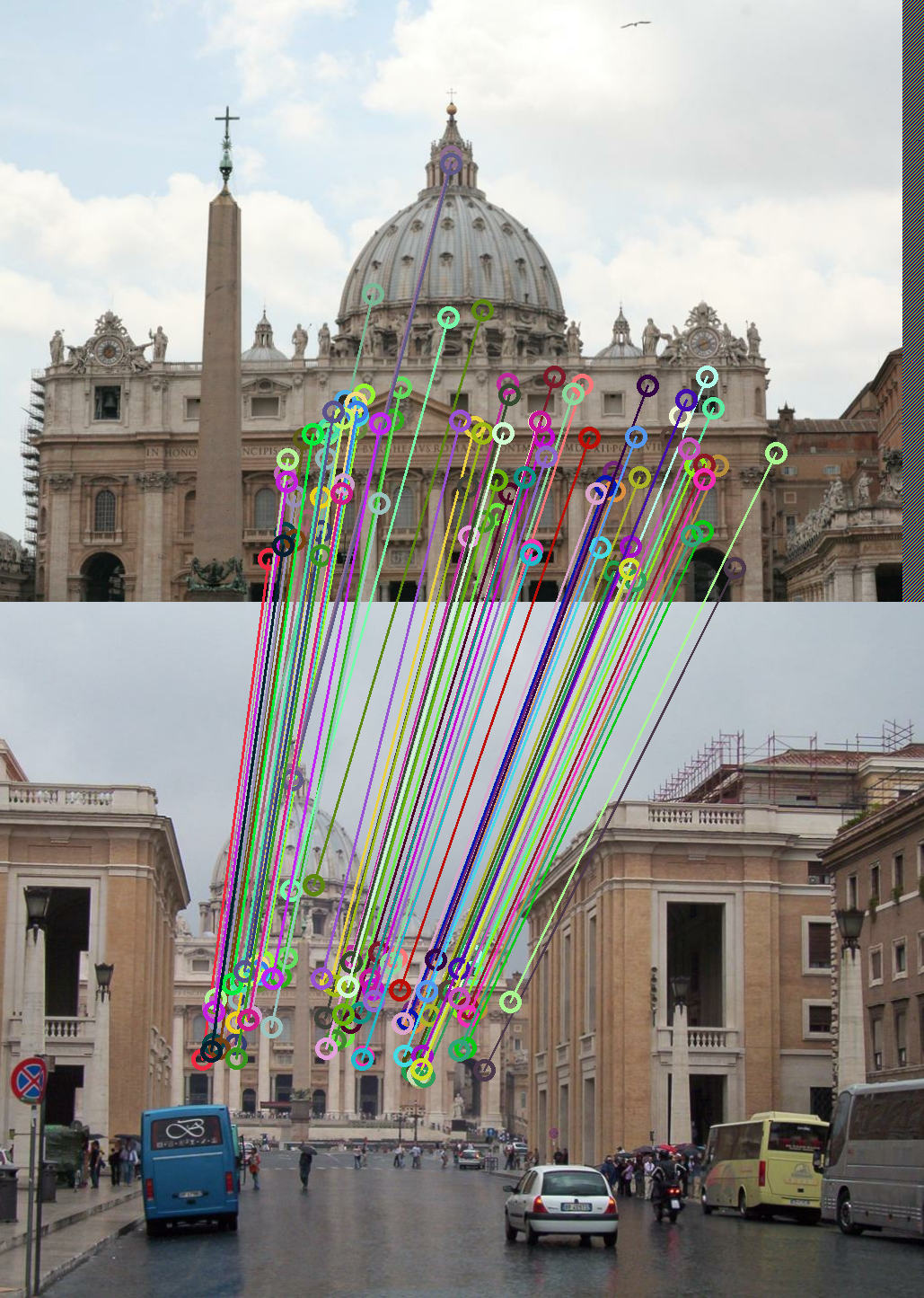}
\caption{St. Peter's Square}
\end{subfigure}
\begin{subfigure}{0.425\textwidth}
\centering
\includegraphics[trim={0mm 0mm 180mm 0mm},clip,width=\textwidth]{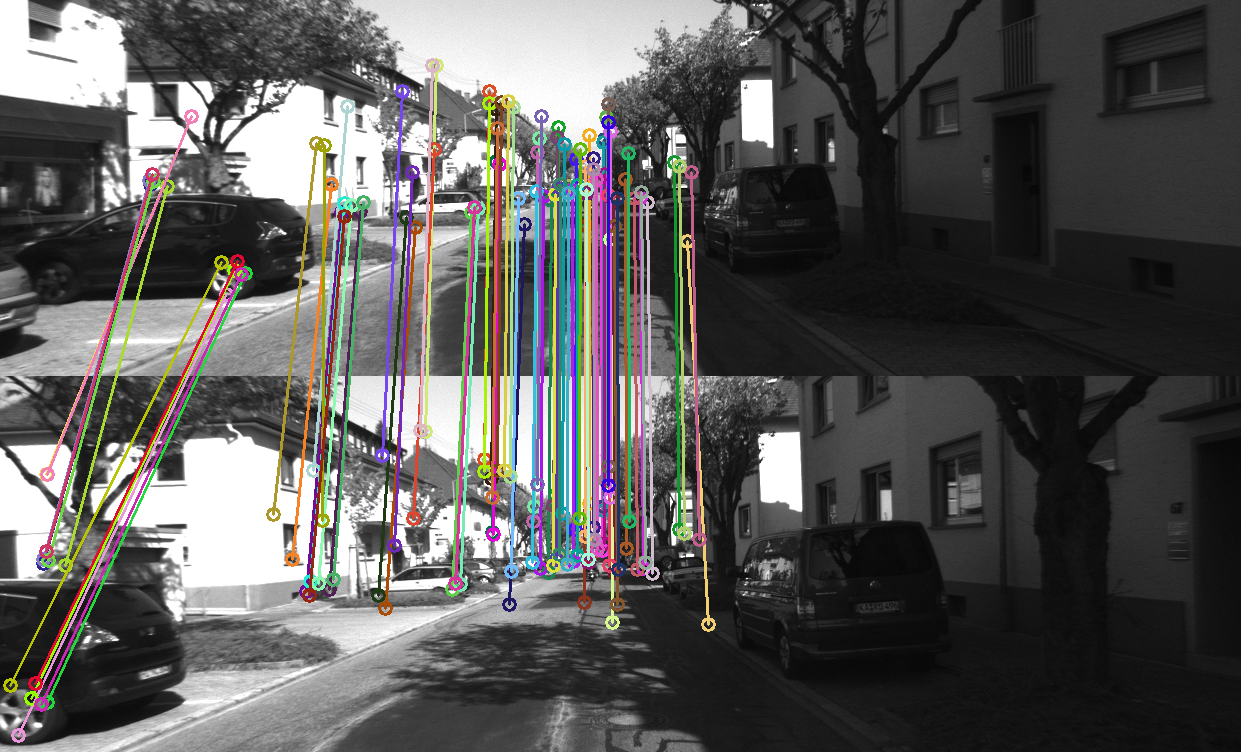}
\caption{KITTI}
\end{subfigure}
\caption{Example image pairs from the PhotoTourism~\cite{cvpr2020ransactutorial} and KITTI~\cite{Geiger2012CVPR} datasets where the proposed SIFT-based solver estimates the (a) fundamental and (b) essential matrix and (c) solves the semi-calibrated case (\ie, unknown focal length). A hundred random inliers are drawn. }
\label{fig:example_pairs}
\end{figure}

Using partially affine co-variant features for model estimation is a known approach with a number of papers published in the recent years. 
In \cite{mills2018four}, the feature orientations are used to estimate the essential matrix. 
In \cite{barath2017phaf}, the fundamental matrix is assumed to be a priori known and an algorithm is proposed for approximating a homography exploiting the rotations and scales of two SIFT correspondences. The approximative nature comes from the assumption that the scales along the axes are equal to the SIFT scale and the shear is zero. In general, these assumptions do not hold. 
The method of Barath et al.~\cite{barath2018approximate} approximates the fundamental matrix by enforcing the geometric constraints of affine correspondences on the epipolar lines. Nevertheless, due to using the same affine model as in \cite{barath2017phaf}, the estimated epipolar geometry is solely an approximation.
\begin{figure}[t]
    \centering
	\begin{subfigure}[t]{0.49\columnwidth}
    	\includegraphics[width = 1.0\columnwidth,trim={0 10cm 0 0},clip]{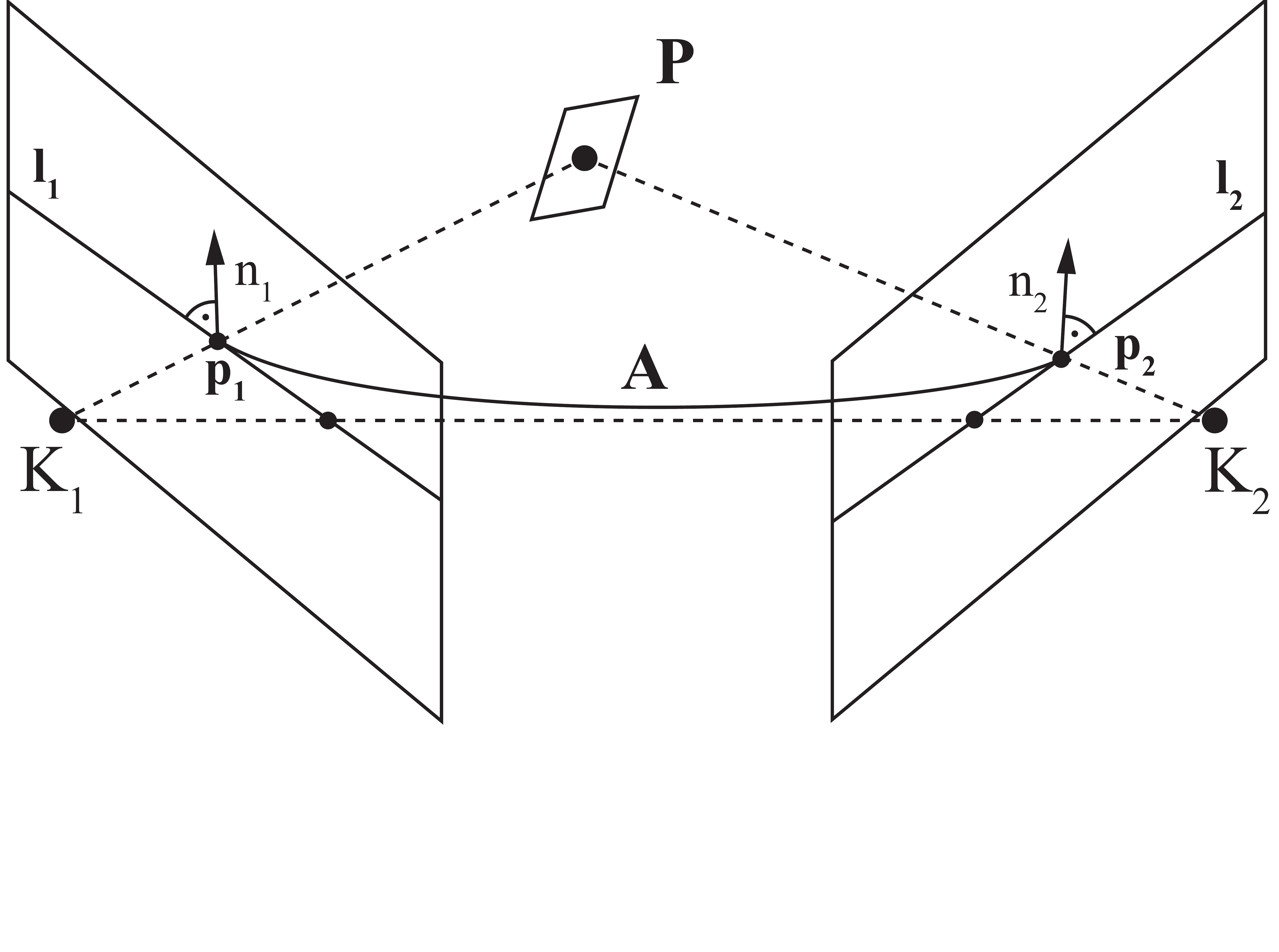}
    	\caption{ Local affine transformation }
		\label{fig:rotating_normals}
	\end{subfigure}\hfill
	\begin{subfigure}[t]{0.49\columnwidth}
    	\includegraphics[width = 1.0\columnwidth,trim={0 10cm 0 0},clip]{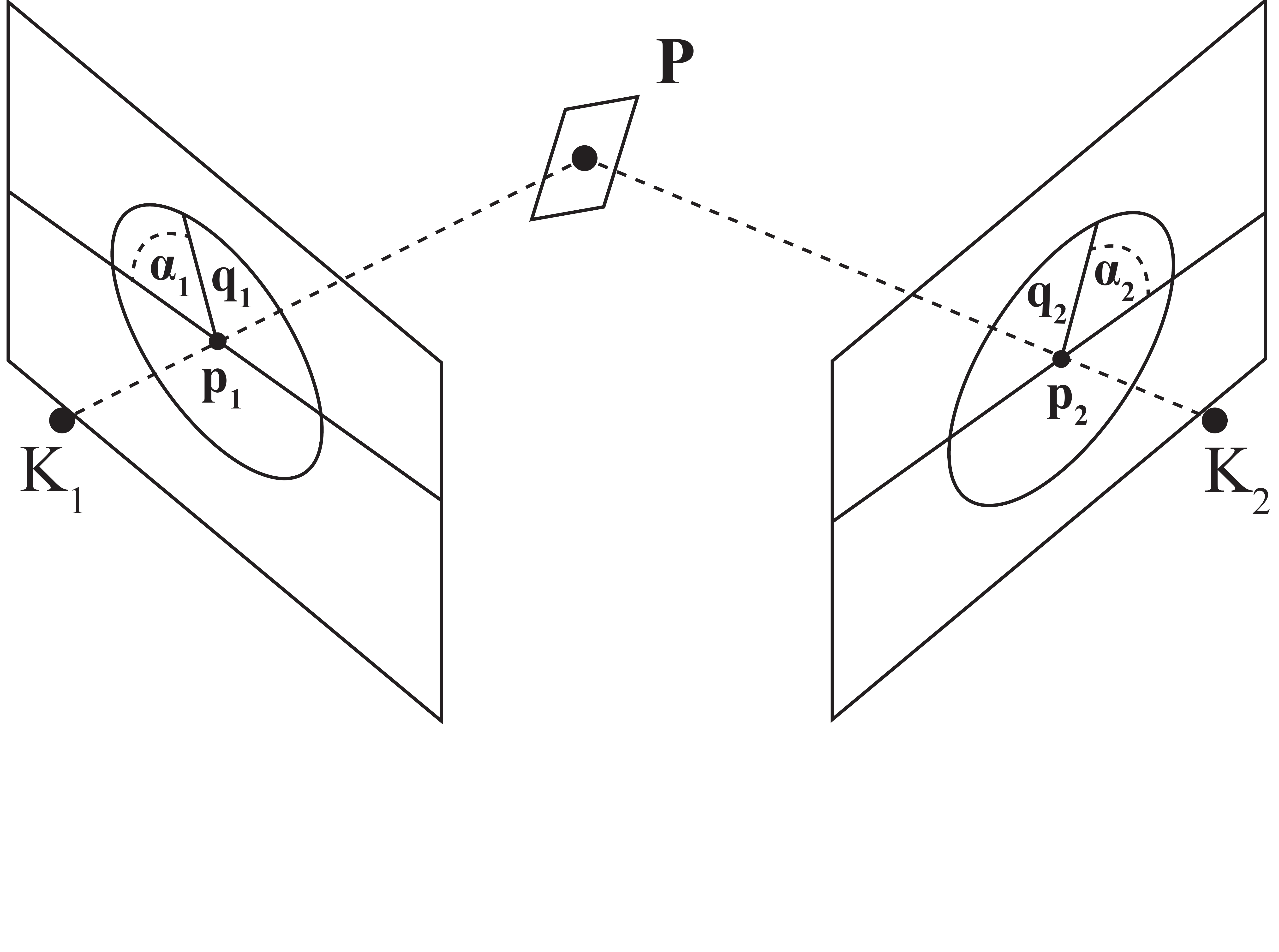}
    	\caption{ Orientation- and scale-covariant features }
		\label{fig:geometric_interpr_sift}
	\end{subfigure}
	\caption{ 
	(a) The geometric interpretation of the relationship of a local affine transformations and the epipolar geometry (Eq.~\eqref{eq:normal_to_normal_2}; proposed in \cite{barath2017focal}). 
	Given the projection $\matr{p}_i$ of point $\matr{P}$ in the $i$th camera $\matr{K}_i$, $i \in \{1,2\}$. 
	The normal $\matr{n}_1$ of epipolar line $\matr{l}_1$ is mapped by affinity $\Aff \in \mathbb{R}^{2 \times 2}$ into the normal $\matr{n}_2$ of epipolar line $\matr{l}_2$.
	(b) Visualization of the orientation- and scale-covariant features. Point $\textbf{P}$ and the surrounding patch projected into cameras $\textbf{K}_1$ and $\textbf{K}_2$. 
	The rotation of the feature in the $i$th image is $\alpha_i \in [0, 2\pi)$ and the size is $q_i \in \mathbb{R}$. The scaling from the $1$st to the $2$nd image is calculated as $q = q_2 / q_1$ and the rotation as $\alpha = \alpha_2 - \alpha_1$. }
\end{figure} 
In \cite{barath2018five}, a two-step procedure is proposed for estimating the epipolar geometry. First, a homography is obtained from three oriented features. Finally, the fundamental matrix is retrieved from the homography and two additional correspondences. Even though this technique considers the scales and shear as unknowns, thus estimating the epipolar geometry instead of approximating it, the proposed decomposition of the affine matrix is not justified theoretically. Therefore, the geometric interpretation of the feature rotations is not provably valid. 
Barath et al.~\cite{barath2018recovering} proposes a way of recovering affine correspondences from the feature rotation, scale, and the fundamental matrix. 
In \cite{barath2019homography}, a method is proposed to estimate the homography from two SIFT correspondences and a theoretically justifiable affine decomposition and general constraints on the homography are  provided. 
Even though having a number of methods estimating geometric entities from SIFT features, there are \textit{no solvers} that exploit the feature orientations and scales for estimating the epipolar geometry in the general case.  
The reason is that the parameterization in \cite{barath2019homography} does not allow directly solving for the relative pose since each new correspondence yields two equations and, also, two additional unknowns -- no constraint is gained on epipolar geometry from considering the orientation and scale.

The contributions of the paper are: (i) 
We introduce new constraints relating the oriented circles centered on the observed point locations. These constraints relate the SIFT orientations and scales in two images with the elements of affine correspondence $\Aff$. 
As such, we show that constraints relating $\Aff$ and the parameters of a SIFT correspondence derived  in~\cite{barath2019homography} do not describe the full geometric relationship and, therefore, are not sufficient for estimating the epipolar geometry. 
(ii) Exploiting the new constraints that relate $\Aff$ and the SIFT correspondence, we derive the geometric relationship between orientation and scale-covariant features and epipolar geometry.
The new SIFT-based constraint is a linear equation that can be straightforwardly used together with the well-known epipolar constraint to efficiently solve relative pose problems.
(iii) Finally, we exploit the new constraint in minimal solvers for estimating epipolar geometry of uncalibrated, calibrated and partially-calibrated cameras with unknown focal length. 
The new solvers require four SIFT correspondences for estimating the fundamental matrix and three for finding the essential matrix both in the fully and in the partially-calibrated cases. 
The reduced sample size accelerates randomized robust estimation by a large margin on a number of real-world datasets while often leading to better accuracy.
Example image pairs are shown in Fig.~\ref{fig:example_pairs}.

\section{Theoretical Background}
\
\noindent
\textbf{Affine correspondence} $(\Point_1, \Point_2, \Aff)$ is a triplet, where $\Point_1 = [u_1 \; v_1 \; 1]^\trans$ and $\Point_2 = [u_2 \; v_2 \; 1]^\trans$ are a corresponding homogeneous point pair in two images and $\Aff$
is a $2 \times 2$ linear transformation which is called \textit{local affine transformation}. Its elements in a row-major order are: $a_1$, $a_2$, $a_3$, and $a_4$. To define $\Aff$, we use the definition provided in~\cite{Molnar2014} as it is given as the first-order Taylor-approximation of the $\text{3D} \to \text{2D}$ projection functions. For perspective cameras, the formula for $\Aff$ is the first-order approximation of the related \textit{homography} matrix
as: 
\begin{eqnarray}
		\begin{array}{lllllll}
          a_{1} & = & \frac{\partial \textbf{u}_2}{\partial u_1} = \frac{h_{1} - h_{7} u_2}{s}, & & 
          a_{2} & = & \frac{\partial \textbf{u}_2}{\partial v_1} = \frac{h_{2} - h_{8} u_2}{s},  \\[2mm]
          a_{3} & = & \frac{\partial \textbf{v}_2}{\partial u_1} = \frac{h_{4} - h_{7} v_2}{s}, & &
          a_{4} & = & \frac{\partial \textbf{v}_2}{\partial v_1} = \frac{h_{5} - h_{8} v_2}{s}, 
		\end{array}
        \label{eq:taylor_approximation}
\end{eqnarray}
where $\textbf{u}_i$ and $\textbf{v}_i$ are coordinate functions given the projection function in the $i$th image ($i \in \{1,2\}$) and $s = u_1 h_7 + v_1 h_8 + h_9$ is the projective depth. The elements of homography $\Hom$ in a row-major order are written as $h_1$, $h_2$, ..., $h_9$.

The relationship of an affine correspondence and a homography is described by six linear equations~\cite{barath2017theory}. Since an affine correspondence contains a point pair, the well-known equations (from $\alpha \Hom \Point_1 = \Point_2$, $\alpha \in \mathbb{R}$) relating the point coordinates hold \cite{hartley2003multiple}. The two equations are written as follows:
\begin{equation}
	\begin{array}{ll}
    \label{eq:orig_dlt}
        u_1 h_1 + v_1 h_2 + h_3 - u_1 u_2 h_7 - v_1 u_2 h_8 - u_2 h_9 &= 0, \\
        u_1 h_4 + v_1 h_5 + h_6 - u_1 v_2 h_7 - v_1 v_2 h_8 - v_2 h_9 &= 0.
    \end{array}
\end{equation}
After re-arranging Eq.~\eqref{eq:taylor_approximation}, 4 linear constraints are obtained from $\Aff$ as
\begin{equation*}
    \small
    \begin{aligned}
    	\label{eq:ha}
    	h_{1} - \left( u_2 + a_{1} u_1  \right) h_{7} - a_{1} v_1 h_{8} - a_{1} h_{9} &= 0, &
    	h_{2} - \left( u_2 + a_{2} v_1  \right) h_{8} - a_{2} u_1 h_{7} - a_{2} h_{9} &= 0, \\
    	h_{4} - \left( v_2 + a_{3} u_1  \right) h_{7} - a_{3} v_1 h_{8} - a_{3} h_{9} &= 0, &
    	h_{5} - \left( v_2 + a_{4} v_1  \right) h_{8} - a_{4} u_1 h_{7} - a_{4} h_{9} &= 0. 
    \end{aligned}
\end{equation*}
Consequently, an affine correspondence provides six linear equations in total, for the elements of the related homography matrix. 

\textbf{Fundamental matrix} $\Fund$
relating two images of a rigid scene is a $3 \times 3$ projective transformation ensuring the so-called epipolar constraint 
\begin{eqnarray}
    \label{eq:epipolar}
    \Point_2^\trans \matr{F} \Point_1 = 0.
\end{eqnarray}
Since its scale is arbitrary and $\det(\Fund) = 0$, matrix $\Fund$ has 7 degrees-of-freedom (DoF). 

The relationship of the epipolar geometry (either a fundamental or essential matrix) and affine correspondences are described in \cite{barath2017focal} through the effect of $\mat A$ on the corresponding epipolar lines. 
Suppose that fundamental matrix $\matr{F}$, point pair $\matr{p}$, $\matr{p}'$, and the related affinity $\matr{A}$ are given. 
It can be proven that $\matr{A}$ transforms $\matr{v}$ to $\matr{v}'$, where $\matr{v}$ and $\matr{v}'$ are the directions of the epipolar lines ($\matr{v}, \matr{v}' \in \mathbb{R}^2$ \emph{s.t.} $\norm{\matr{v}}_2 = \norm{\matr{v}'}_2 = 1$) in the first and second images \cite{bentolila2014conic}, respectively. 
It can be seen that transforming the infinitesimally close vicinity of $\matr{p}$ to that of $\matr{p}'$, $\matr{A}$ has to map the lines going through the points.
Therefore, constraint $\matr{A} \matr{v} \parallel \matr{v}'$ holds. 

As it is well-known~\cite{turkowski1990transformations}, formula $\matr{A} \matr{v} \parallel \matr{v}'$ can be reformulated as follows: 
\begin{equation}
	\label{eq:normal_to_normal_1}
	\matr{A}^{-\trans} \matr{n} = \beta \matr{n}',
\end{equation} 
where $\matr{n}$ and $\matr{n}'$ are the normals of the epipolar lines ($\matr{n}, \matr{n}' \in \mathbb{R}^2$ \emph{s.t.} $\matr{n} \bot \matr{v}$, $\matr{n'} \bot \matr{v'}$). 
Scalar $\beta$ denotes the scale between the transformed and the original vectors if $\norm{\matr{n}}_2 = \norm{\matr{n}'}_2 = 1$. 
The normals are calculated as the first two coordinates of epipolar lines
\begin{equation}
	\matr{l} = \matr{F}^\trans \matr{p}' = \left[ a \; b \; c \right]^\trans, \quad \matr{l}' = \matr{F} \matr{p} = \left[ a' \; b' \; c' \right]^\trans.
	\label{eq:epi_lines}
\end{equation}
Since the common scale of normals $\matr{n} = \matr{l}_{[1:2]} = \left[ a \; b \right]^\trans$ and $\matr{n}' = \matr{l}_{[1:2]}' = \left[ a' \; b' \right]^\trans$ comes from the fundamental matrix, Eq.~\eqref{eq:normal_to_normal_1} is modified as follows:
\begin{equation}
	\label{eq:normal_to_normal_2}
	\matr{A}^{-\trans} \matr{n} = -\matr{n}'.
\end{equation} 
Formulas \eqref{eq:epi_lines} and \eqref{eq:normal_to_normal_2} yield two equations which are linear in the parameters of the fundamental matrix as:
\begin{eqnarray}
	\label{eq:affine_constraint_1}
	\begin{array}{r}
	(u' + a_1 u) f_1 + a_1 v f_2 + a_1 f_3 + (v' + a_3 u) f_4 + 
    a_3 v f_5 + a_3 f_6 + f_7 = 0, 
    \end{array}\\
    \label{eq:affine_constraint_2}
    \begin{array}{r}
    	a_2 u f_1 + (u' + a_2 v) f_2 + a_2 f_3 + a_4 u f_4 + 
        (v' + a_4 v) f_5 + a_4 f_6 + f_8 = 0. 
    \end{array}
\end{eqnarray}
%
Points ($u_1$, $v_1$) and ($u_2$, $v_2$) are the points in, respectively, the first and second images.
%

In summary, \textit{the linear part} of a local affine transformation \textit{gives two linear equations}, Eqs.~\eqref{eq:affine_constraint_1} and \eqref{eq:affine_constraint_2}, for epipolar geometry estimation. 
A point correspondence yields a third one, Eq.~\eqref{eq:epipolar}, through the epipolar constraint. 
Thus, an affine correspondence yields three linear constraints. As the fundamental matrix has seven DoF, three affine correspondences are enough for estimating $\matr F$ \cite{barath2020making}.\footnote{Precisely, fundamental matrix $\matr F$ can be estimated from two affine and a point correspondence.}
Essential matrix $\Ess$ has five DoF and, thus, two affine correspondences are enough for the estimation \cite{barath2018efficient}.

\section{Epipolar Geometry and SIFT Features}
\label{sec:upgrading}

In this section, we show the relationship between epipolar geometry and orientation and scale-covariant features.
Even though we will use SIFT as an alias for this kind of features, the derived formulas hold for all of them.
First, the affine transformation model is described in order to interpret the SIFT angles and scales.
This model is substituted into the relationship of affine transformations and epipolar geometry.
Combining the derived constraint via Gr\"obner-basis, we finally propose a linear equation characterizing the epipolar consistency of the orientation and scale part of the SIFT features.


\subsection{Interpretation of SIFT Features}

Reflecting the fact that we are given a scale $q_i \in \mathbb{R}^+$ and rotation $\alpha_i \in [0, 2 \pi)$ independently in each image ($i \in \{ 1, 2 \}$; see Fig.~\ref{fig:geometric_interpr_sift}), the objective is to define affine correspondence $\Aff$ as a function of them. 
Such an interpretation was proposed in~\cite{barath2019homography}.
In this section, we simplify the formulas in \cite{barath2019homography} in order to reduce the number of unknowns in the system.
%
%
To understand the orientation and scale part of SIFT features, we exploit the definition of affine correspondences proposed by Barath et al.~\cite{barath2015optimal}. In~\cite{barath2015optimal}, $\Aff$ is defined as the multiplication of the Jacobians of the projection functions w.r.t.\ the image directions in the two images as follows:
\begin{equation}
    \textbf{A} = \matr{J}_2 \matr{J}_1^{-1},
    \label{eq:A_as_jacobians}
\end{equation}
where $\matr{J}_1$ and $\matr{J}_2$ are the Jacobians of the 3D $\to$ 2D projection functions. Proof is in~\cite{barath2019homography}.
For the $i$th Jacobian, we use the following decomposition:
\begin{equation}
    \small
    \textbf{J}_i = \textbf{R}_i \textbf{U}_i = \begin{bmatrix} 
        \cos(\alpha_i) & -\sin(\alpha_i) \\
        \sin(\alpha_i) & \cos(\alpha_i)
    \end{bmatrix} 
    \begin{bmatrix} 
        q_{u,i} & w_i \\
        0 & q_{v,i}
    \end{bmatrix},
    \label{eq:J_model}
\end{equation}
where angle $\alpha_i$ is the rotation in the $i$th image, $q_{u,i}$ and $q_{v,i}$ are the scales along axes $u$ and $v$, and $w_i$ is the shear. 
Plugging Eq.~\eqref{eq:J_model} into Eq.~\eqref{eq:A_as_jacobians} leads to
%
    $\textbf{A} = \textbf{R}_2 \textbf{U}_2 (\textbf{R}_1 \textbf{U}_1)^{-1} = \textbf{R}_2 \textbf{U}_2 \textbf{U}_1^{-1} \textbf{R}_1^{\trans}$,
%
where $\textbf{U}_1$ and $\textbf{U}_2$ contain the unknown scales and shears in the two images. 
Since we are not interested in finding them separately, we replace $\textbf{U}_2 \textbf{U}_1^{-1}$ by upper-triangular matrix $\mat U = \textbf{U}_2 \textbf{U}_1^{-1}$ simplifying the formula to
\begin{eqnarray*}
    \textbf{A} = \textbf{R}_2 \mat U \textbf{R}_1^{\trans} = 
    \begin{bmatrix} 
        \cos(\alpha_2) & -\sin(\alpha_2) \\
        \sin(\alpha_2) & \cos(\alpha_2)
    \end{bmatrix}
    \begin{bmatrix} 
        q_{u} & w \\
        0 & q_{v}
    \end{bmatrix}
    \begin{bmatrix} 
        \cos(\alpha_1) & \sin(\alpha_1) \\
        -\sin(\alpha_1) & \cos(\alpha_1)
    \end{bmatrix}. 
\end{eqnarray*}
Angles $\alpha_1$ and $\alpha_2$ are known from the SIFT features. 
Let us use notation $c_i = \cos(\alpha_i)$ and $s_i = \sin(\alpha_i)$.
The equations after the matrix multiplication become
\begin{eqnarray*}
    \small
    \textbf{A} =
    \begin{bmatrix}
        a_1 & a_2 \\
        a_3 & a_4
    \end{bmatrix} = 
    \begin{bmatrix} 
        c_2 (c_1 q_{u} - s_1 w) + s_2 s_1 q_v & c_2 (s_1 q_u + c_1 w) - s_2 c_1 q_{v} \\
        s_2 (c_1 q_{u} - s_1 w) - c_2 s_1 q_v & s_2 (s_1 q_u + c_1 w) + c_2 c_1 q_{v}
    \end{bmatrix}. 
\end{eqnarray*}
After simplifying the equations, we get the following linear system
\begin{eqnarray}
    \begin{array}{lrclr}
    a_1 & = c_2 c_1 q_{u} - c_2 s_1 w + s_2 s_1 q_v, & \quad & 
    a_2 & = c_2 s_1 q_u + c_2 c_1 w - s_2 c_1 q_{v},\\
    a_3 & = s_2 c_1 q_{u} - s_2 s_1 w - c_2 s_1 q_v, & \quad &
    a_4 & = s_2 s_1 q_u + s_2 c_1 w + c_2 c_1 q_{v},
    \end{array}
    \label{eq:DecomposedParameters}
\end{eqnarray}
where the unknowns are the affine parameters $a_1$, $a_2$, $a_3$, $a_4$, scales $q_u$, $q_v$ and shear $w$.

In addition to the previously described constraints, we are given two additional ones. 
First, it can be seen that the uniform scales of the SIFT features are proportional to the area of the underlying image region and, therefore, the scale change provides constraint 
\begin{equation}
    \small
    \det \textbf{A} = \det \left(\textbf{R}_2 \mat U \textbf{R}_1^{\trans} \right) = \det \mat U = q_u q_v = \frac{q_2^2}{q_1^2},
    \label{eq:ScaleDependency}
\end{equation}
where $q_1$ and $q_2$ are the SIFT scales in the two images. 
Second, the SIFT orientations and scales in the two images provide an additional constraint as
\begin{equation}
    \small
    q_1 \textbf{A} 
    \begin{bmatrix}
        \cos(\alpha_1) \\
        \sin(\alpha_1)
    \end{bmatrix} = q_2 \begin{bmatrix}
        \cos(\alpha_2) \\
        \sin(\alpha_2)
    \end{bmatrix}
    \label{eq:CircleConstraint}
\end{equation}
relating the oriented circles centered on the point correspondence.
Next, we show how these constraints can be used to derive the constraint relating the SIFT orientation and scale with the epipolar geometry.

\subsection{SIFT Epipolar Constraint}
\label{sec:Sift_constraint}

Our goal is to derive a constraint that relates epipolar geometry and orientation- and scale-covariant features. To do this, we first consider the constraints that relate the elements of $\matr{A}$ and the measured orientations $\alpha_i$ and scales $q_i$
of the features in the first and second images. In~\cite{barath2019homography}, such constraints were derived by eliminating $q_u$, $q_v$ and $w$ from the ideal generated by~\eqref{eq:DecomposedParameters},~\eqref{eq:ScaleDependency} and trigonometric identities $c_i^2 + s_i^2 = 1$ for $i \in \{1,2\}$, using the elimination ideal technique~\cite{kukelovaCVPR2017}. This method resulted into two constraint, \ie, the generators of the elimination ideal, one of which is directly~\eqref{eq:ScaleDependency} and the second one has the form 
 \begin{eqnarray}
    \small
    \label{eq:constraint_old}
    c_1 s_2 a_1 + s_1 s_2 a_2 - c_1 c_2 a_3 - c_2 s_1 a_4 = 0.
\end{eqnarray}
 Here, we will show 
 that once the constraints~\eqref{eq:CircleConstraint} are added to the ideal,  and we ensure $q_1 \neq 0$ and $q_2 \neq 0$ by saturating the ideal with $q_1$ and $q_2$, then the elimination ideal is generated directly by constraints~\eqref{eq:ScaleDependency} and~\eqref{eq:CircleConstraint}. This means that for the derivation of the constraints that relate the elements of matrix 
  $\matr{A}$ and the measured orientations $\alpha_i$ and scales $q_i$,  equations~\eqref{eq:DecomposedParameters} are not necessary. 
  These new constraints have the following form
   \begin{eqnarray}
      \small
    \label{eq:constraint1}
    a_2 a_3 - a_1 a_4 + q^2 &=& 0,\\
    \label{eq:constraint2}
    a_3 c_1+a_4 s_1 - s_2 q &=& 0, \\
    \label{eq:constraint3}
    a_1 c_1+a_2 s_1 - c_2 q &=& 0,
\end{eqnarray}
where   $q= \frac{q_2}{q_1}$.
  Moreover, thanks to constraints~\eqref{eq:CircleConstraint} relating the oriented circles centered on the points and which were not used in~\cite{barath2019homography}, 
  we have three constraints, compared to the two polynomials derived in~\cite{barath2019homography}\footnote{Note that the constraint 
~\eqref{eq:constraint_old} derived in~\cite{barath2019homography} is a linear combination of constraints~\eqref{eq:constraint2} and~\eqref{eq:constraint3} and can be obtained by eliminating $q$ from these two equations.}.
This will help us to derive a new constraint relating epipolar geometry and covariant features that was not possible to derive using only the two constraints proposed in~\cite{barath2019homography}. For this purpose, we 
  create an ideal $J$ generated by polynomials~\eqref{eq:constraint1}-\eqref{eq:constraint3},~\eqref{eq:affine_constraint_1} and~\eqref{eq:affine_constraint_2}. 
    Then the unknown elements of the affine transformation $\Aff$ are eliminated from the generators of $J$.
We do this by computing the generators of 
the elimination ideal
$J_1 = J \cap \mathbb{C}[f_1,\dots,f_9,u_1,v_1,u_2,v_2,q,s_1,c_1,s_2,c_2]$.
The elimination ideal $J_1$ is generated by polynomial
\begin{eqnarray}
\label{eq:constraint_F}
c_2qf_1u_1+s_2qf_4u_1+c_2qf_2v_1+s_2qf_5v_1+c_2qf_3+s_2qf_6+\\
\nonumber
c_1f_1u_2+s_1f_2u_2+c_1f_4v_2+s_1f_5v_2+c_1f_7+s_1f_8= 0.
\end{eqnarray}
Note that~\eqref{eq:constraint_F} is linear in the elements of $\Fund$ and, as such, it can be straightforwardly used together with the well-known epipolar constraint for point correspondences to estimate the epipolar geometry.

\subsection{Solvers for Epipolar Geometry}
\label{sec:ess_solver}
In this section, we will describe different solvers for estimating epipolar geometry using orientation- and scale-covariant features (\eg, SIFT correspondences). In Section~\ref{sec:Sift_constraint}, we showed that each SIFT correspondence gives us two linear constraints on the elements of the fundamental (or essential) matrix. 
One constraint is the well-known epipolar constraint~\eqref{eq:epipolar} for point correspondences and one is the new derived SIFT-based constraint~\eqref{eq:constraint_F}.
As such, we can directly transform \textit{all} existing point-based solvers for estimating epipolar geometry to solvers working with SIFT features. 
The only difference will be that for solvers that estimate the geometry from $n$ point correspondences, we will use $\lceil \frac{n}{2} \rceil$ SIFT ones, and in the solver we will replace $\lfloor \frac{n}{2} \rfloor$ epipolar constraints~\eqref{eq:epipolar} from point correspondences with $\lfloor \frac{n}{2} \rfloor$ SIFT constraints of the form~\eqref{eq:constraint_F}. This will affect only some coefficients in coefficient matrices used in these solvers and not the structure of the solver.
Moreover, for problems where $n$, which in this case corresponds to 
the DoF of the problem, is not a multiple of two, we can use all $\lceil \frac{n}{2} \rceil$ available constraints of the form~\eqref{eq:constraint_F} to simplify the solver. Next, we will describe solutions to three important relative pose problems, \ie for uncalibrated, calibrated, and partially calibrated perspective cameras with unknown focal length. However, note, that our method is not only applicable to these problems and presented solvers, but can be directly applied to all existing point-based solvers for estimating epipolar geometry.

\textbf{Fundamental matrix.} 
This is a 7 DoF problem, which means that we need four SIFT correspondences  $(\matr{p}_1^i, \matr{p}_2^i,\alpha_1^i,\alpha_2^i,q)$, $i \in \{1,2,3,4\}$ to solve it. 
For the $i$th correspondence, the  epipolar constraint~\eqref{eq:epipolar} and the proposed SIFT-based constraint~\eqref{eq:constraint_F}
can be written as $\matr{C}_i \matr{f} = 0$, where 
matrix $\matr{C}_i \in \mathbb{R}^{2\times9}$ is the coefficient matrix consisting of two rows and
vector $\matr{f} = [ f_1,  f_2,  f_3,  f_4,  f_5,  f_6,  f_7,  f_8,  f_9 ]^\trans$ consists of the unknown elements of the fundamental matrix. 
As mentioned above, in this case, we can either use all four constraints of the form~\eqref{eq:constraint_F} and simplify the solver by not using the $\det \mat F = 0$ constraint\footnote{This solver corresponds to the well-known eight-point solver~\cite{hartley2003multiple}}, or we can use just three equations of the form~\eqref{eq:constraint_F}  and solve the obtained cubic polynomial implied by the constraint $\det \mat F = 0$. In our experiments, we decided to test the second solver, which corresponds to the well-known seven-point solver~\cite{hartley2003multiple} and which leads to more accurate results.


\textbf{Essential matrix.}
The relative pose problem for calibrated cameras is a 5 DoF problem and we need three SIFT correspondences  $(\matr{p}_1^i, \matr{p}_2^i,\alpha_1^i,\alpha_2^i,q)$, $i \in \{1,2,3\}$ to solve it. 
Similarly to the uncalibrated case, for the $i$th correspondence, the epipolar constraint and the new SIFT-based one
can be written as $\matr{C}_i \matr{e} = 0$, where 
$\matr{e} = [ e_1,  \dots,  e_9 ]^\trans$ 
is the vector of the unknown elements of the essential matrix. 
Matrix $\matr{C}_i \in \mathbb{R}^{2\times9}$ is the coefficient matrix consisting of two rows, the first one containing coefficients from the epipolar constraint and the second one from the SIFT-based one~\eqref{eq:constraint_F}.
Considering the three feature case, $\matr{C}$ is of size $6 \times 9$ as $\matr{C} = [ \matr{C}_1^\trans, \matr{C}_2^\trans, \matr{C}_3^\trans ]^\trans$.
While using the top $5 \times 9$ sub-matrix of $\matr{C}$ would allow using the well-known solvers for solving the five-point problem~\cite{nister2004efficient,li2006five,hartley2012efficient}, having $6$ rows in $\matr{C}$ to use simpler solvers.
We, thus, adopt the solver from~\cite{barath2018efficient} proposed, originally, for estimating from affine correspondences.

First, the $3$-dimensional null-space of $\matr{C}$ is obtained by, \eg, LU decomposition as it is significantly faster than the SVD and Eigen decompositions.
The solution is given by the linear combination of the three null vectors as
	$\matr{x} = \alpha \matr{n}_1 + \beta \matr{n}_2 + \gamma \matr{n}_3$,
%
where $\matr{n}_1$, $\matr{n}_2$, and $\matr{n}_3$ are the null vectors and  parameters $\alpha$, $\beta$, and $\gamma$ are unknown non-zero scalars. These scalars are defined up to a common scale, therefore, one of them can be chosen to an arbitrary value. In the proposed algorithm, $\gamma = 1$. 

By substituting this formula to the trace constraint, \ie, $\matr{E} \matr{E}^\trans \matr{E} - \frac{1}{2} \text{trace}(\matr{E} \matr{E}^\trans) \matr{E} = 0$) and the determinant constraint $\det \matr{E} = 0$, ten polynomial equations are given. 
They can be formed as $\matr{Q} \matr{y} = \matr{b}$, where $\matr{Q}$ and $\matr{b}$ are the coefficient matrix and the inhomogeneous part (\ie, coefficients of monomial $1$), respectively. Vector $\matr{y} = [ \alpha^3 ,  \beta^3 ,  \alpha^2 \beta ,  \alpha \beta^2 ,  \alpha^2 ,  \beta^2 ,  \alpha \beta ,  \alpha ,  \beta ]^\trans$ consists of the monomials of the system. 
Matrix $\matr{Q}$ is of size $10 \times 9$, therefore, the system is over-determined since ten equations are given for nine unknowns. 
Its optimal solution in least squares sense is given by $\matr{y} = \matr{Q}^{\dag} \matr{b}$, where matrix $\matr{Q}^{\dag}$ is the Moore-Penrose pseudo-inverse of matrix $\matr{Q}$. 
The solver has only a single solution which is beneficial for the robust estimation.

The elements of the solution vector $\matr{y}$ are dependent. Thus $\alpha$ and $\beta$ can be obtained in multiple ways, \eg, as $\alpha_1 = y_8$, $\beta_1 = y_9$ or $\alpha_2 = \sqrt[3]{y_1}$, $\beta_2 = \sqrt[3]{y_2}$. 
To choose the best candidates, we paired every possible $\alpha$ and $\beta$ and selected the one minimizing the trace constraint $\matr{E} \matr{E}^\trans \matr{E} - \frac{1}{2} \text{trace}(\matr{E} \matr{E}^\trans) \matr{E} = 0$. 

\textbf{Fundamental matrix and focal length.} 
Assuming the unknown common focal length in both cameras, the relative pose problem has 6 DoF. As such, it can be solved from three SIFT correspondences $(\matr{p}_1^i, \matr{p}_2^i,\alpha_1^i,\alpha_2^i,q)$, $i \in \{1,2,3\}$. In this case, three SIFT correspondences generate exactly the minimal case.
We can apply one of the standard 6PT solvers~\cite{li2006simple,stewenius2008minimal,hartley2012efficient,kukelovaCVPR2017}. We choose the method from \cite{kukelovaCVPR2017} that uses elimination ideals to eliminate the unknown focal length and generates a smaller elimination template matrix than the original Gr\"{o}bner basis solver~\cite{stewenius2008minimal}. 

\section{Experiments}

In this section, we test the proposed SIFT-based solvers in a fully controlled synthetic environment and on a number of publicly available real-world datasets. 

\subsection{Synthetic Experiments}

\begin{figure}[t!]
	\begin{center}
	\begin{subfigure}[t]{0.325\columnwidth}
		\includegraphics[width=1.0\columnwidth,trim={1mm 0mm 10mm 1mm},clip]{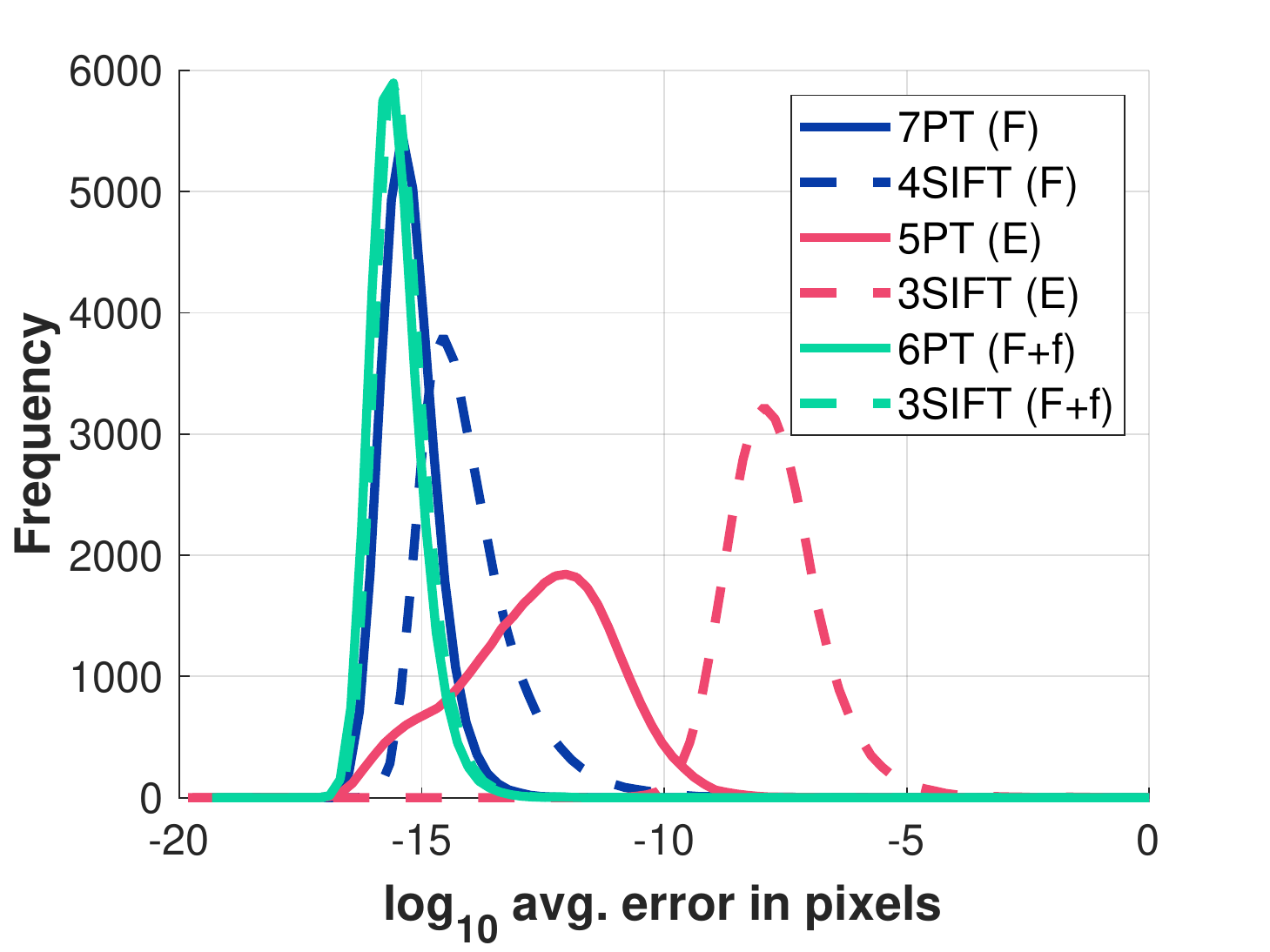}
		\caption{}
		\label{fig:stability}
	\end{subfigure}
	\begin{subfigure}[t]{0.325\columnwidth}
		\includegraphics[width=1.0\columnwidth,trim={1mm 0mm 10mm 1mm},clip]{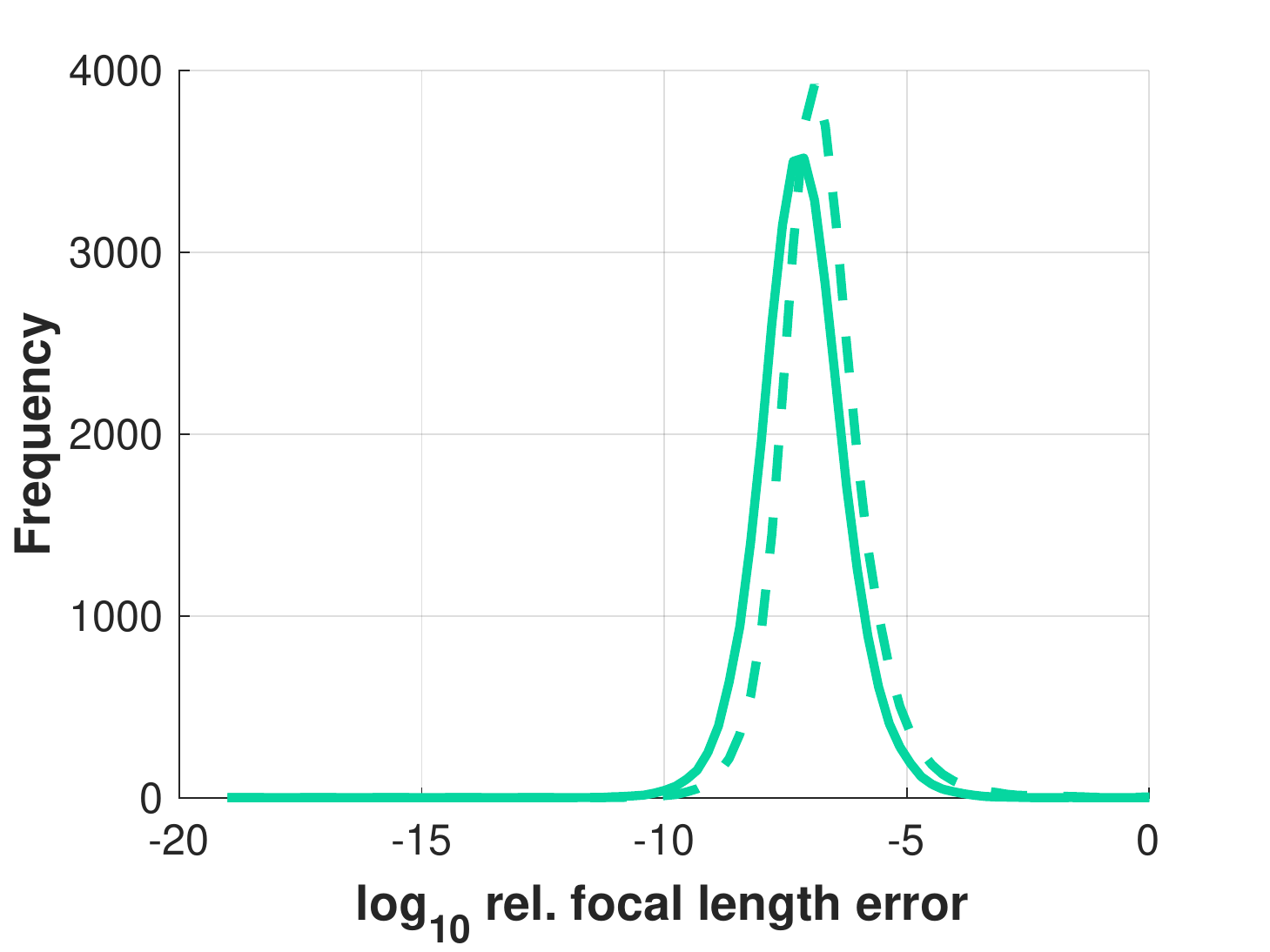}
		\caption{}
		\label{fig:stability_focal}
	\end{subfigure}
	\begin{subfigure}[t]{0.325\columnwidth}
		\includegraphics[width=1.0\columnwidth,trim={1mm 0mm 10mm 1mm},clip]{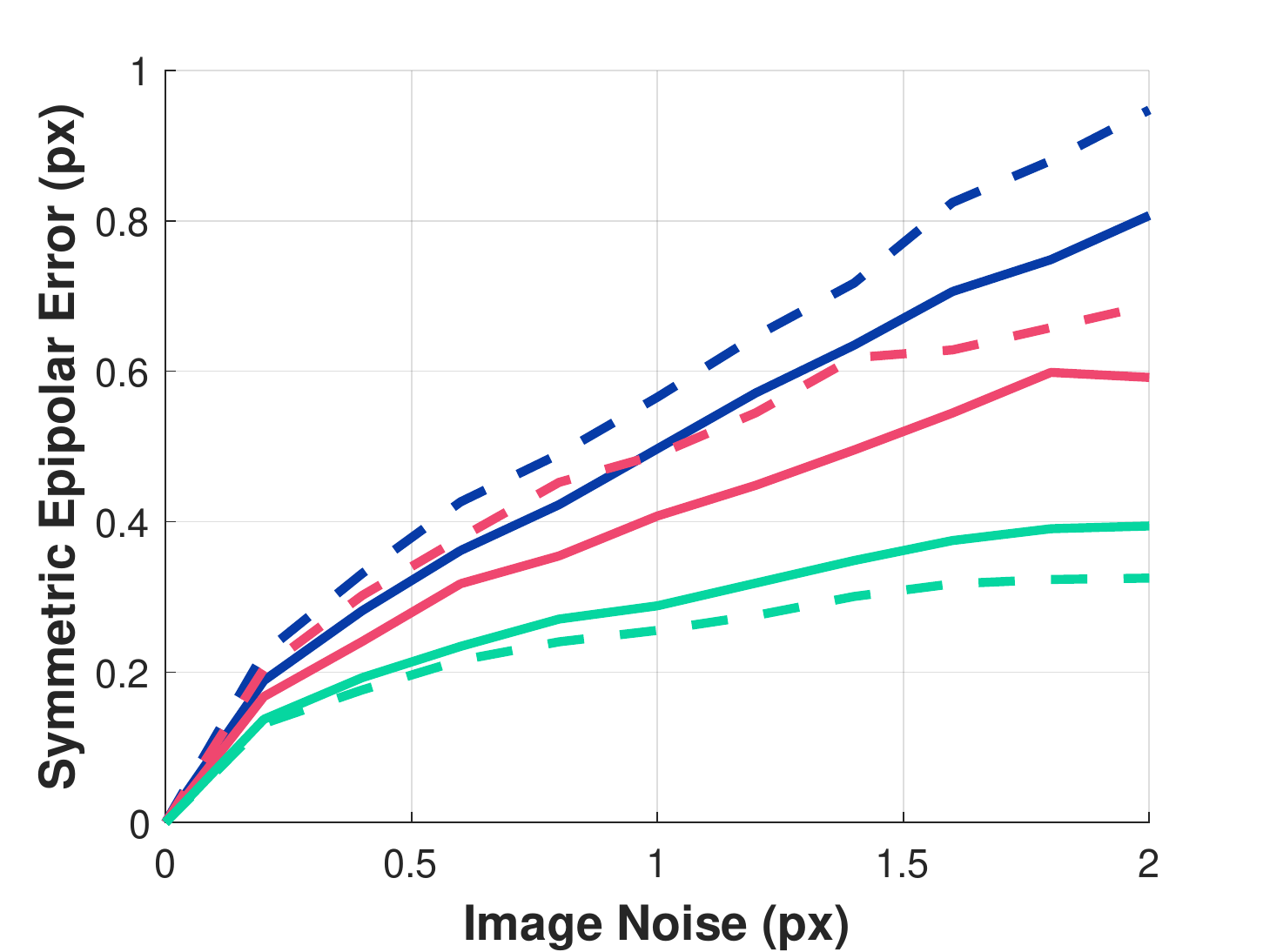}
		\caption{}
		\label{fig:image_noise}
	\end{subfigure}
	\end{center}
	\caption{ \textit{Synthetic experiments.} (a) The frequencies (\num{100000} runs; vertical axis) of $\text{log}_{10}$ sym.\ epipolar errors (horizontal; in pixels) in the essential and fundamental matrices estimated by point and SIFT-based solvers. 
	(b) The frequencies of $\text{log}_{10}$ relative focal length errors (horizontal) estimated by point and SIFT-based solvers. 
	(c) The symmetric epipolar error plotted as a function of the image noise in pixels. 
	}
\end{figure}

To test the accuracy of the relative pose obtained by exploiting the proposed SIFT constraint, first, we created a synthetic scene consisting of two cameras represented by their $3 \times 4$ projection matrices $\matr{P}_1$ and $\matr{P}_2$. 
They were located randomly on a center-aligned sphere with its radius selected uniformly randomly from range $[0.1, 10]$. 
Two planes with random normals were generated at most one unit far from the origin. For each plane, ten random points, lying on the plane, were projected into both cameras.
Note that we need the correspondences to originate from at least two planes in order to avoid having a degenerate situation for fundamental matrix estimation.
To get the ground truth affine transformation for a correspondence originating from the $j$th plane, $j \in \{1, 2\}$, we calculated homography $\Hom_j$ by projecting four random points from the plane to the cameras and applying the normalized DLT~\cite{hartley2003multiple} algorithm. 
The local affine transformation of each correspondence was computed from the ground truth homography by (\ref{eq:taylor_approximation}). 
Note that $\Hom$ could have been calculated directly from the plane parameters. 
However, using four points promised an indirect but geometrically interpretable way of noising the affine parameters: adding noise to the coordinates of the four points initializing $\Hom$. 
To simulate the SIFT orientations and scales, $\Aff$ was decomposed to $\matr{J}_1$, $\matr{J}_2$. 
Since the decomposition is ambiguous, $\alpha_1$, $q_{u,1}$, $q_{v,1}$, $w_1$ were set to random values. $\matr{J}_1$ was calculated from them. Finally, $\matr{J}_2$ was calculated as $\matr{J}_2 = \Aff \matr{J}_1$.
Zero-mean Gaussian-noise was added to the point coordinates, and, also, to the coordinates which were used to estimate the affine transformations.

Fig.~\ref{fig:stability} reports the numerical stability of the methods in the noise-free case. The frequencies (vertical axis), \ie, the number of occurrences in \num{100000} runs, are plotted as the function of the $\log_{10}$ average symmetric epipolar error (in pixels; horizontal) computed from the estimated model and the unused correspondences.
All methods on all problems lead to stable solutions. 
While the 3SIFT essential matrix solver seems the least stable, it is important to note that the horizontal axis is in pixels and, therefore, having ${\approx}10^{-5}$ pixel maximum error can be considered stable. 
Fig.~\ref{fig:stability_focal} reports the numerical stability of the estimated focal lengths in the semi-calibrated case.
The horizontal axis is the $\log_{10}$ relative focal length error calculated as $\epsilon_f = |f_{\text{est}} - f_{\text{gt}}| / f_{\text{gt}}$.
Both methods lead to stable solutions.

In Figure~\ref{fig:image_noise}, 
the average symmetric epipolar (over \num{10000} runs; in pixels) errors are plotted as the function of the image noise added both to the point coordinates and affine parameters (indirectly, via contaminating the initializing homography).
The error is calculated on correspondences not used for the estimation.
The SIFT-based solvers are slightly more sensitive to the noise than the point-based one. 
This is expected since the image noise has a larger impact on the affine parameters, due to their localized nature, than on the point coordinates~\cite{barath2020making}.
Interestingly, this is not the case when solving the semi-calibrated case, where the SIFT-based solver leads to the most accurate relative poses.
Still, the main message from Figure~\ref{fig:image_noise}
is that the solvers behave reasonably well against increasing image noise. 
In the next section, we will show that, due to the reduced combinatorics of the problem, the SIFT-based methods often yield more accurate solutions than their point-based counterparts inside RANSAC.

\subsection{Real-world Experiments}

\begin{table}[t]
\centering
\setlength{\tabcolsep}{4pt}\setlength\aboverulesep{0pt}\setlength\belowrulesep{0pt}%
\resizebox{1.0\columnwidth}{!}{\begin{tabular}{  l |  c c c c c | c c c c c  }
 	\hline
 		 \rowcolor{black!10}
 	     & \multicolumn{5}{c|}{KITTI (\num{69537} image pairs)} & \multicolumn{5}{c}{PhotoTourism  (\num{9900} image pairs)} \\
 	\hline
 		 Solver & $\epsilon_\textbf{R}$ ($^\circ$) & $\epsilon_\textbf{t}$ ($^\circ$) & $\epsilon_f$ & $t$ (ms) & \# iters  & $\epsilon_\textbf{R}$ ($^\circ$) & $\epsilon_\textbf{t}$ ($^\circ$) & $\epsilon_f$ & $t$ (ms) & \# iters    \\
 	\hline
 		 \rowcolor{black!5}
 		 {\small SIFT-based \textbf{E}} & \textbf{2.8} & 2.2 & -- & \textbf{\phantom{1}53.6} & \textbf{\phantom{1}166} & \textbf{1.3} & 2.3 & -- & \textbf{108.1} & \textbf{2182}  \\
 		 {\small Point-based \textbf{E}} & \textbf{2.8} & \textbf{2.1} & -- & 276.4 & \phantom{1}589 & \textbf{1.3} & \textbf{2.2} & -- & 847.3 & 5059 \\
 	\hline
 		 \rowcolor{black!5}
 		 {\small SIFT-based \textbf{F}} & \textbf{2.7} & \textbf{2.2} & -- & \textbf{\phantom{1}67.3} & \textbf{\phantom{1}304} & \textbf{2.1} & \textbf{6.7} & -- & \textbf{\phantom{1}48.8} & \textbf{4189} \\
 		 {\small Point-based \textbf{F}} & \textbf{2.7} & 2.3 & -- & 154.4 & 1860 & 2.3 & 7.8 & -- & 127.3 & 7145 \\
 	\hline
 		 \rowcolor{black!5}
 		 {\small SIFT-based \textbf{F} + $f$} & \textbf{2.8} & \textbf{2.2} & \textbf{0.77} & \textbf{\phantom{1}61.5} & \textbf{\phantom{1}100} & \textbf{1.5} & \textbf{2.6} & \textbf{0.61} & \textbf{290.7} & \textbf{2386} \\
 		 {\small Point-based \textbf{F} + $f$} & \textbf{2.8} & \textbf{2.2} & 0.80 & 225.8 & \phantom{1}731 & 2.6 & 4.5 & 0.62 & 743.0 & 6423 \\
 	\hline
\end{tabular} }\vspace{1mm}
\caption{ Average rotation, translation (in degrees) and focal length errors, run-times (in milliseconds), and iteration numbers on the KITTI~\cite{Geiger2012CVPR} and PhotoTourism~\cite{cvpr2020ransactutorial} datasets for essential (\textbf{E}) and fundamental (\textbf{F}) matrix estimation and, also, focal length plus fundamental matrix estimation (\textbf{F} + $f$). On the PhotoTourism dataset, we show the median errors. }
\label{table:results}
\end{table}

\begin{figure}[t]
	\begin{center}
	\includegraphics[height=3.5mm]{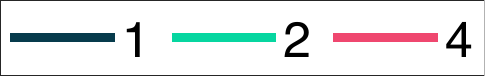}\hspace{4mm}\includegraphics[height=3.5mm]{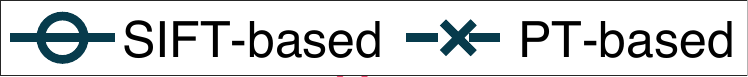}\\[1mm]
	\begin{subfigure}[t]{1.0\columnwidth}
	    \centering
		\includegraphics[width=0.30\columnwidth,trim={1mm 1mm 10mm 1mm},clip]{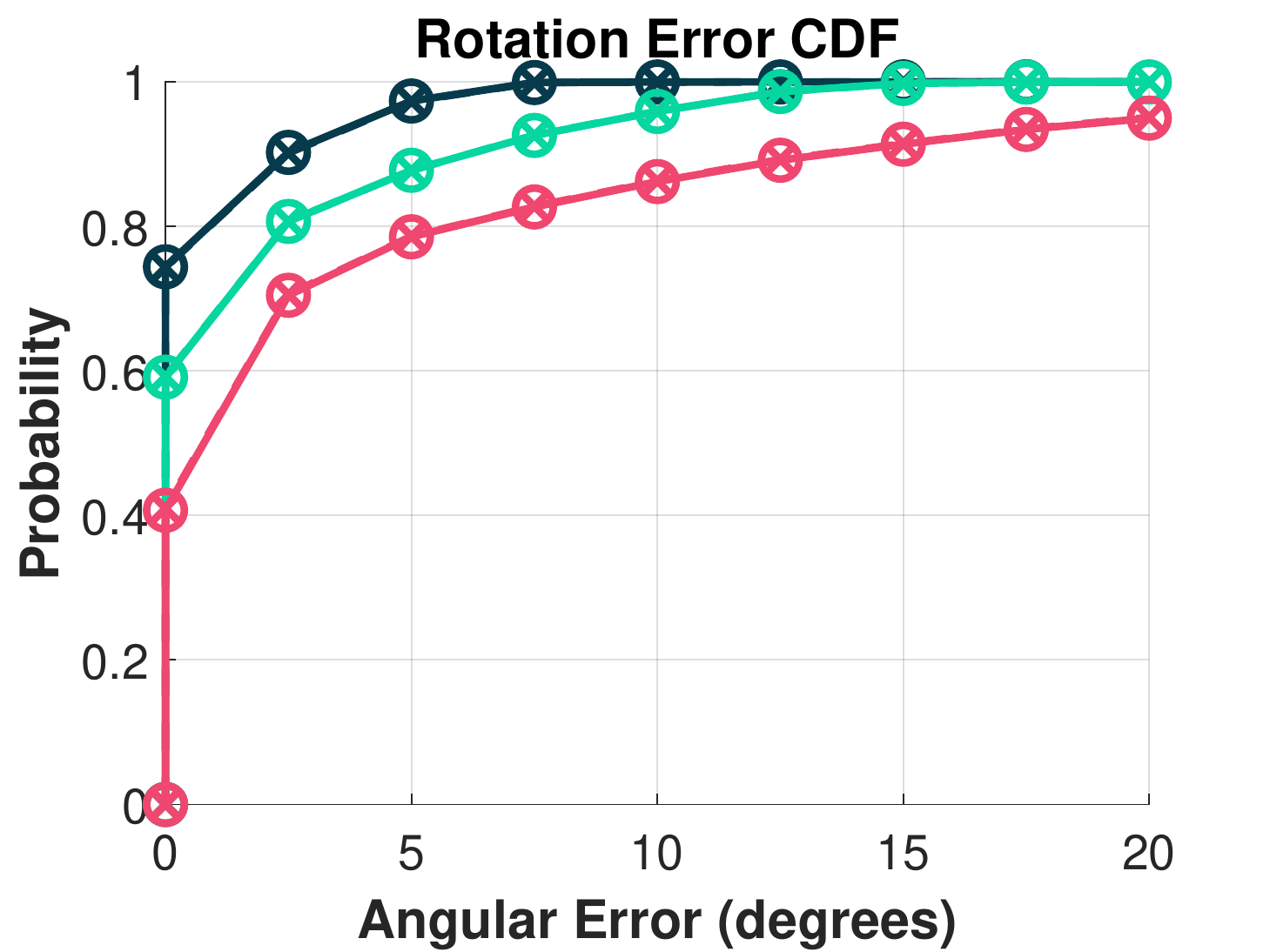}
		\includegraphics[width=0.30\columnwidth,trim={1mm 1mm 10mm 1mm},clip]{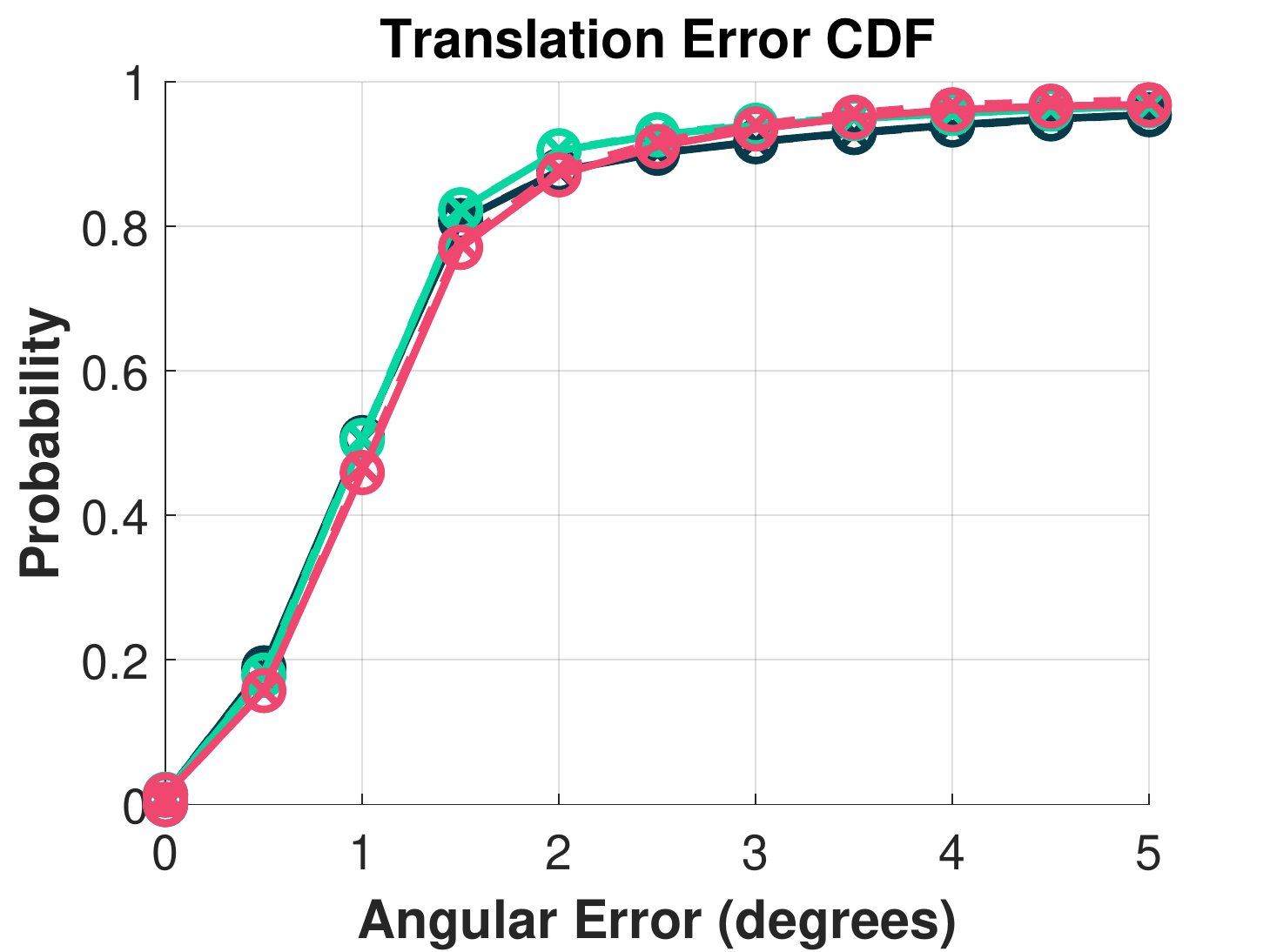}
		\includegraphics[width=0.30\columnwidth,trim={1mm 1mm 10mm 1mm},clip]{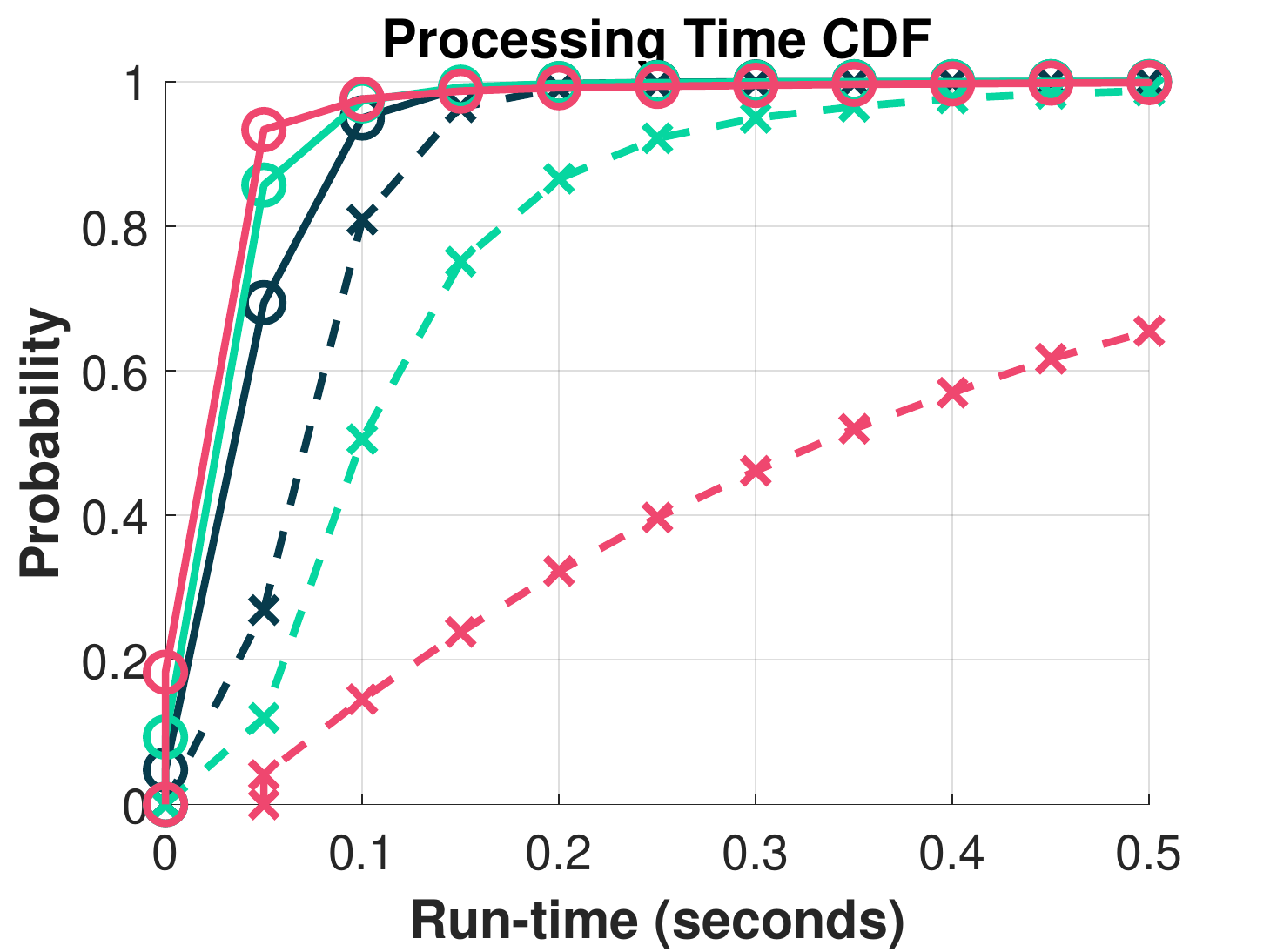}
		\caption{Essential matrix estimation}
		\label{fig:ess_kitti_cdf}
	\end{subfigure}
	\begin{subfigure}[t]{1.0\columnwidth}
	    \centering
		\includegraphics[width=0.30\columnwidth,trim={1mm 1mm 10mm 1mm},clip]{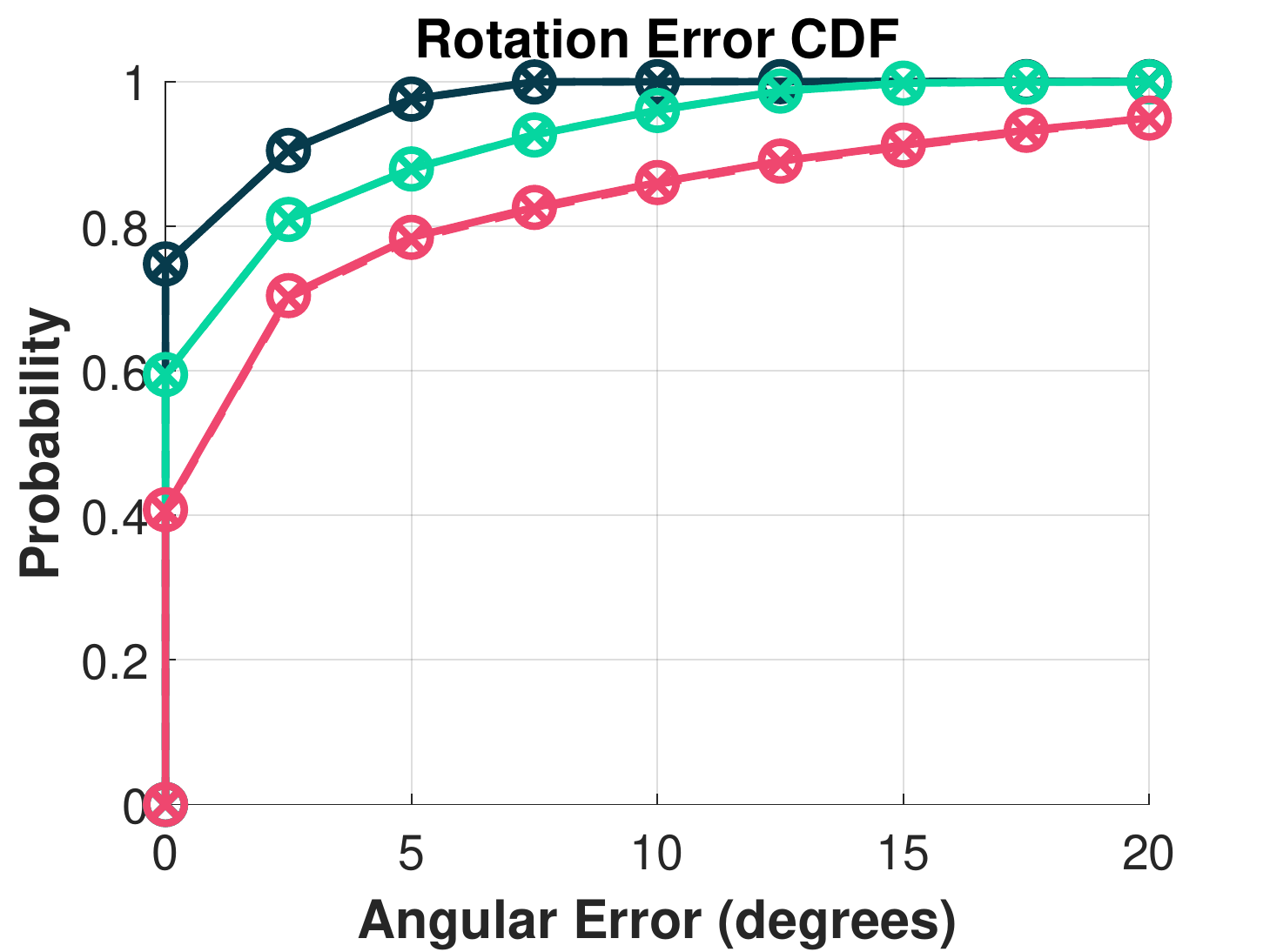}
		\includegraphics[width=0.30\columnwidth,trim={1mm 1mm 10mm 1mm},clip]{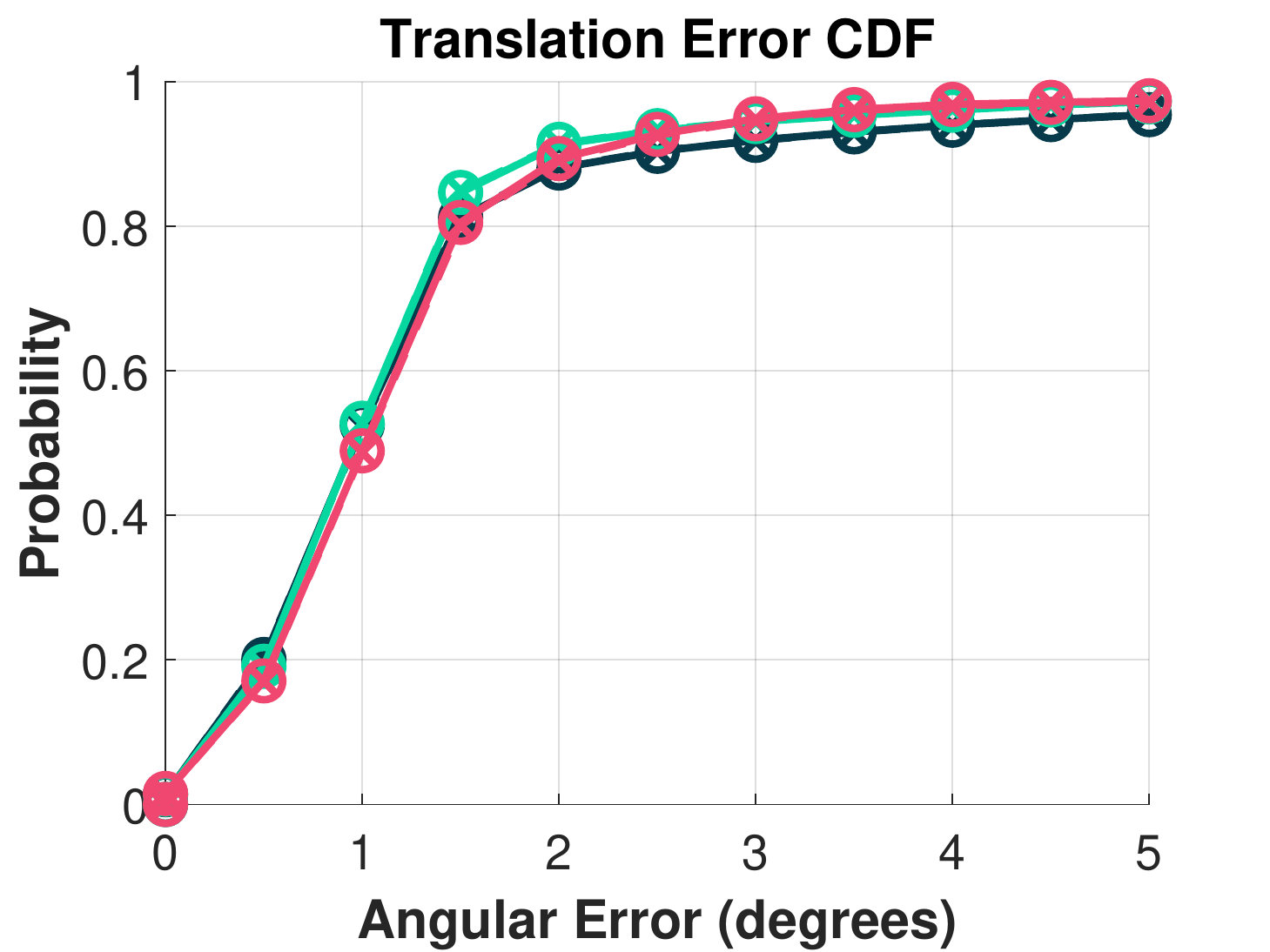}
		\includegraphics[width=0.30\columnwidth,trim={1mm 1mm 10mm 1mm},clip]{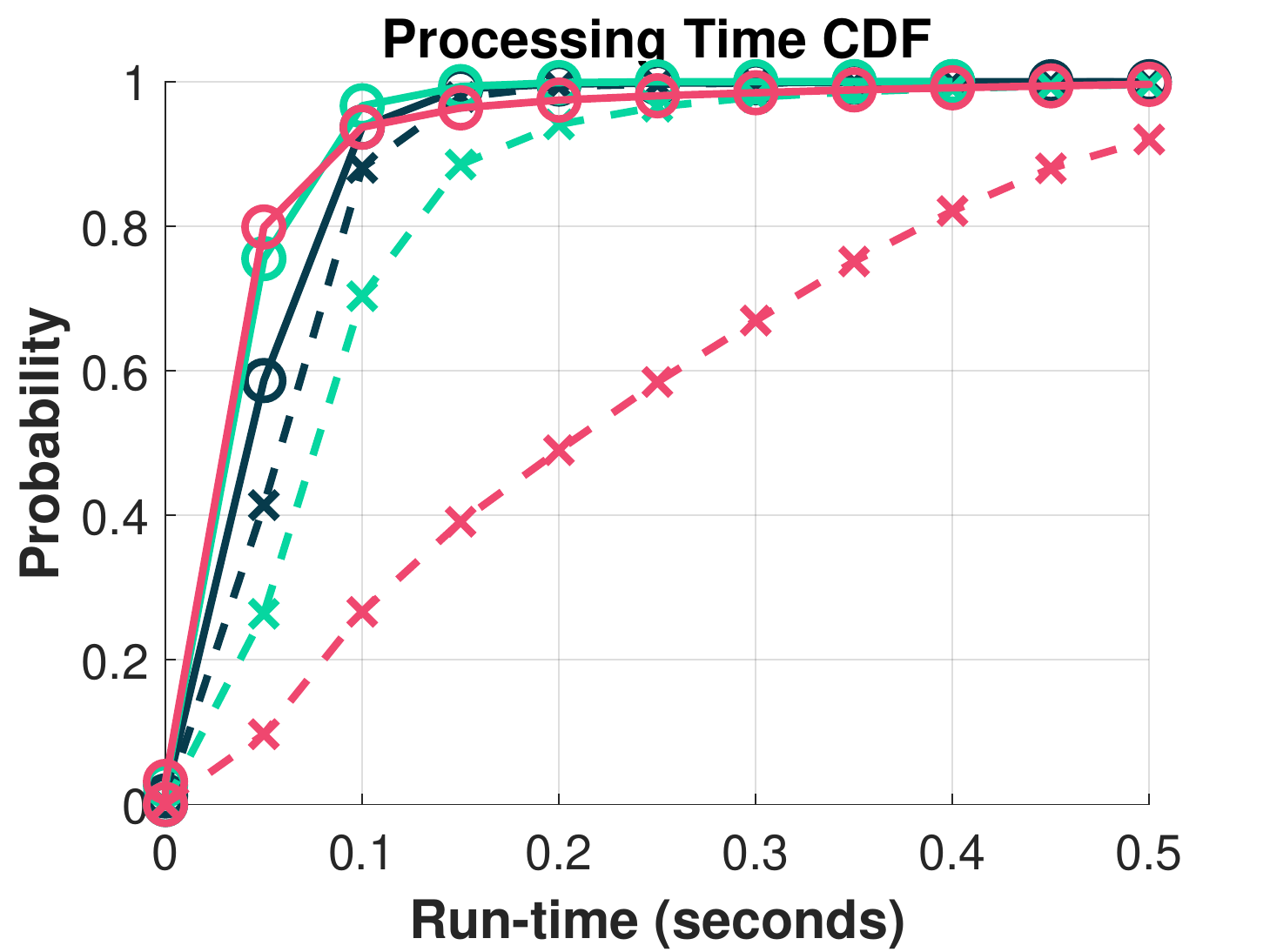}
		\caption{Fundamental matrix estimation}
		\label{fig:fund_kitti_cdf}
	\end{subfigure}
	\begin{subfigure}[t]{1.0\columnwidth}
	    \centering
		\includegraphics[width=0.30\columnwidth,trim={1mm 1mm 10mm 1mm},clip]{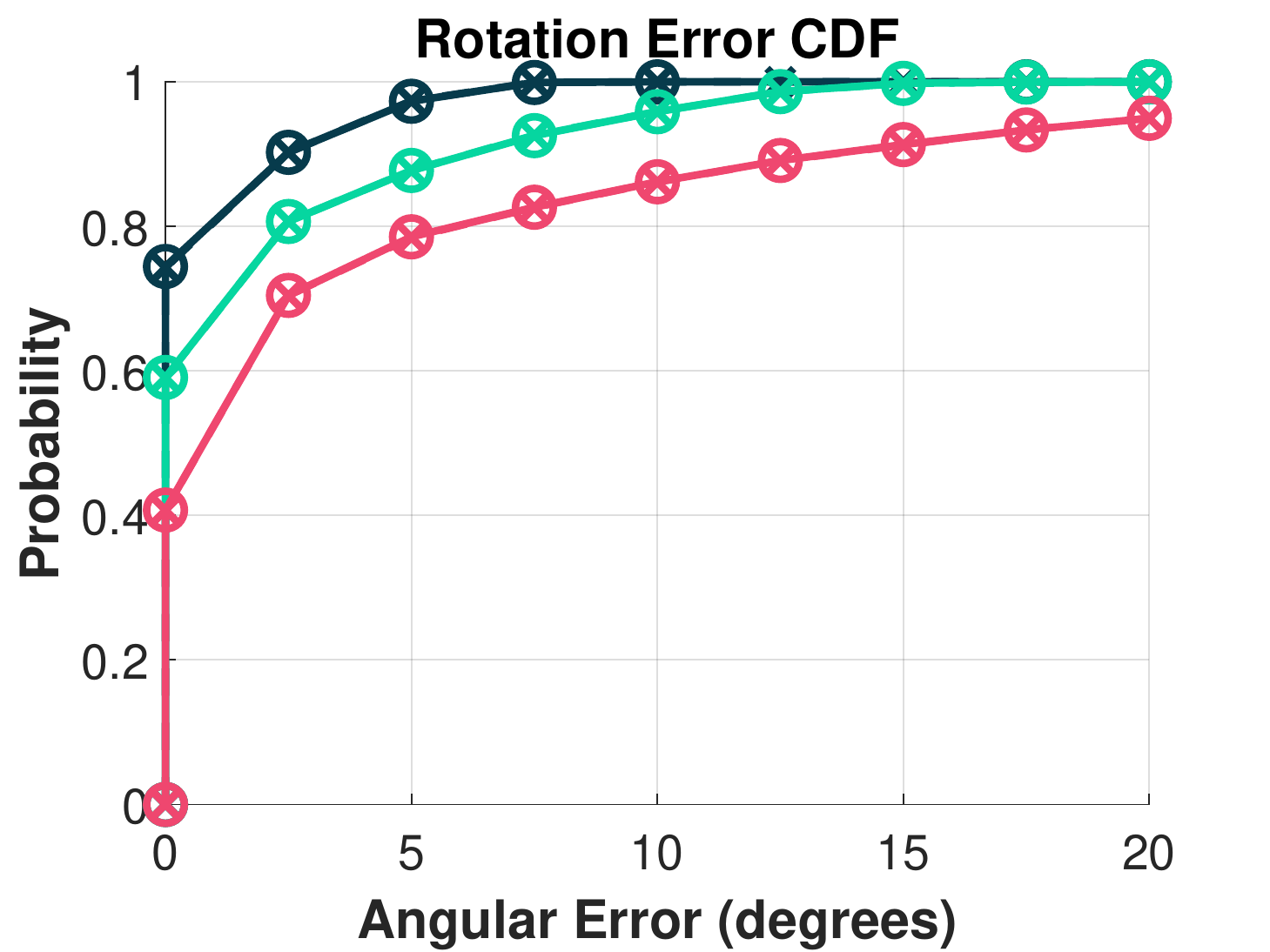}
		\includegraphics[width=0.30\columnwidth,trim={1mm 1mm 10mm 1mm},clip]{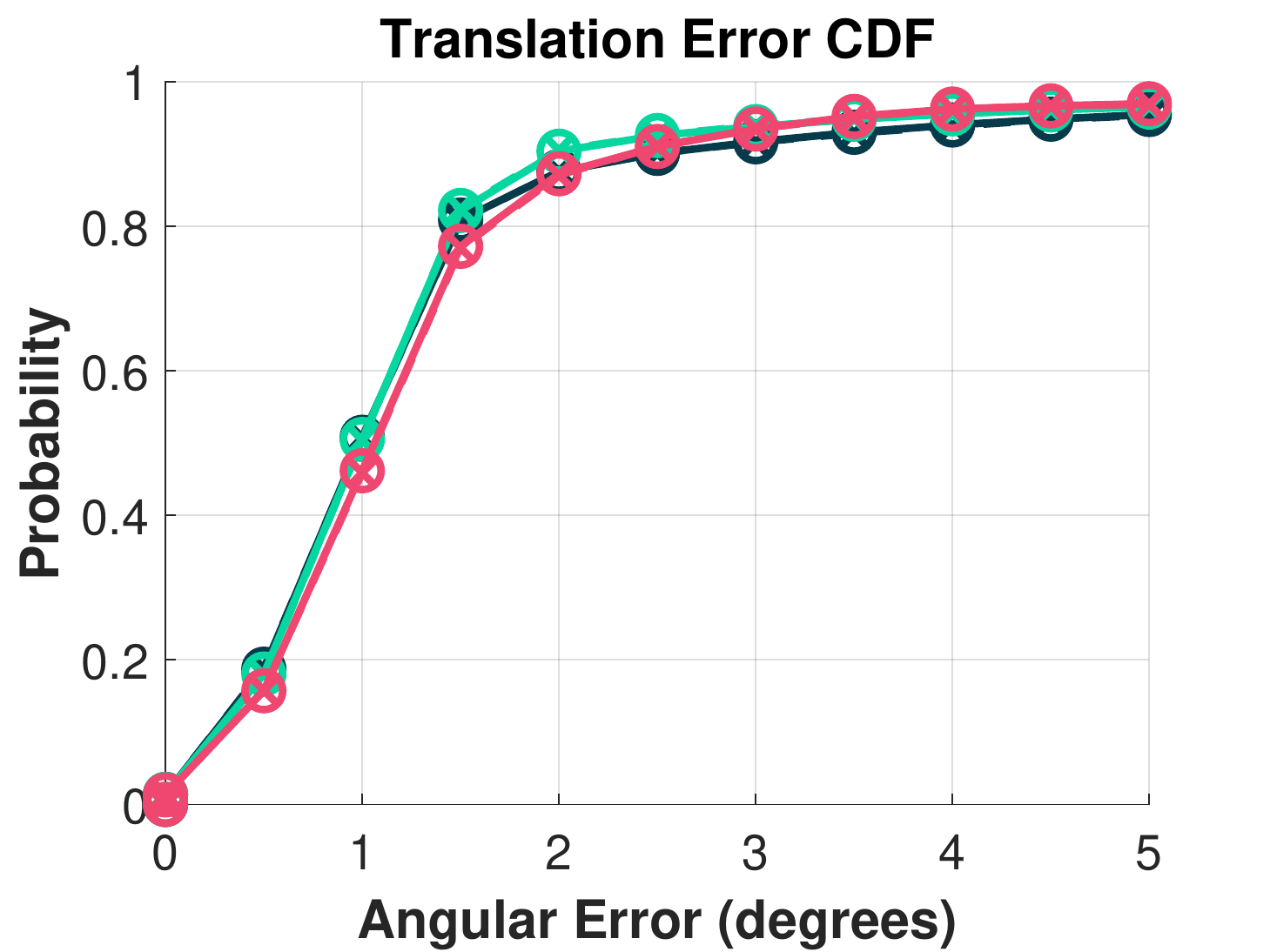}
		\includegraphics[width=0.30\columnwidth,trim={1mm 1mm 10mm 1mm},clip]{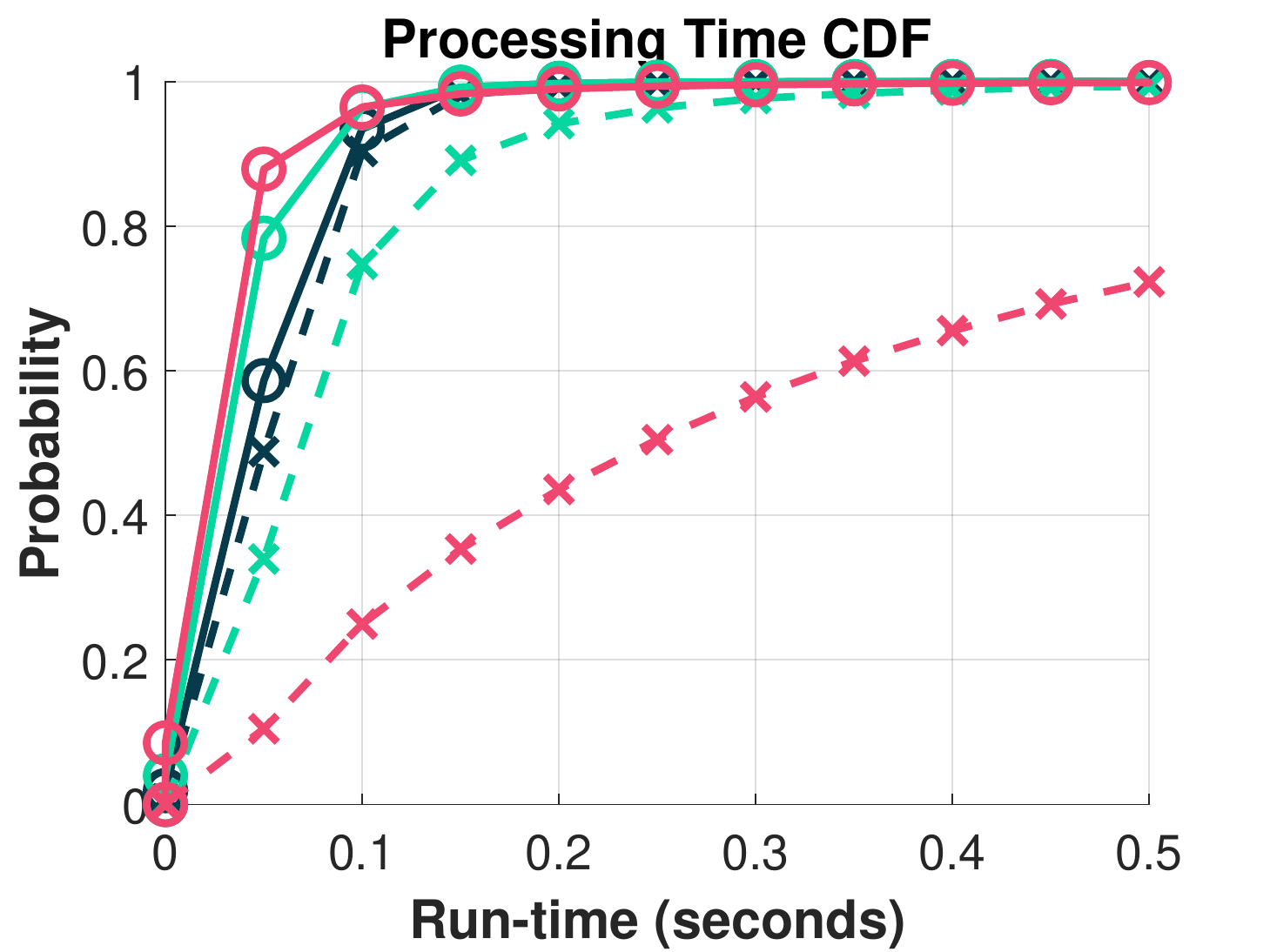}
		\caption{Fundamental matrix and focal length estimation}
		\label{fig:fFf_kitti_cdf}
	\end{subfigure}
	\end{center}
	\caption{
    The cumulative distribution functions of the rotation and translation errors ($^\circ$) and run-times (secs) of epipolar geometry estimation by GC-RANSAC~\cite{barath2017graph} combined with point-based and the proposed SIFT-based minimal solvers on \num{69537} image pairs from the KITTI dataset~\cite{Geiger2012CVPR}.
    The frame difference is denoted by color, \eg, pairs ($I_i$, $I_{i+2}$) are considered for the green curve. }
\end{figure}

For testing the methods, we use the KITTI benchmark~\cite{Geiger2012CVPR} and the datasets from CVPR tutorial \textit{RANSAC in 2020}~\cite{cvpr2020ransactutorial}.
Considering that the orientation and scale of local features are noisier than their point coordinates, we chose to use a locally optimized RANSAC, \ie, GC-RANSAC~\cite{barath2017graph}, as the robust estimator, where the local optimization is applied to only the point coordinates, similarly as in~\cite{barath2019homography,barath2020making}. 
The required confidence is set to $0.99$ and the maximum iteration number to $5000$.

In GC-RANSAC (and other RANSAC-like methods), two different solvers are used: (a) one for fitting to a minimal sample and (b) one for fitting to a non-minimal sample when doing model polishing on all inliers or in the local optimization step. 
For (a), the main objective is to solve the problem using as few points as possible since the run-time depends exponentially on the number of points required for the model estimation. 
The proposed and compared solvers were included in this part of the robust estimator.
%
%

\noindent
\textbf{The KITTI odometry benchmark}
consists of 22 stereo sequences. 
Only 11 sequences (00--10) are provided with ground truth trajectories for training. 
We use these 11 sequences to evaluate the compared solvers. 
Each image is of resolution $1241 \times 376$.
We ran the methods on image pairs formed such that the frame distance is $1$, $2$ or $4$. 
For example, frame distance $2$ means that we form pairs from images $I_i$ and $I_{i+2}$, where $i \in [1, n]$ and $n \in \mathbb{N}^+$ is the number of images in a sequence.
In total, the algorithms were tested on \num{69537} pairs.
To form tentative correspondences, we detected \num{8000} SIFT keypoints in both images to have a reasonably dense point cloud reconstruction and precise camera poses~\cite{IMC2020}.
We combined mutual nearest neighbor check with standard distance ratio test~\cite{lowe1999object} to establish tentative correspondences, as recommended in~\cite{IMC2020}.

\noindent
\textbf{The RANSAC tutorial dataset} comes from the train and validation sets of the CVPR IMW 2020 PhotoTourism challenge.
We use the two scenes, each consisting of \num{4950} image pairs, provided for validation purposes to test the proposed SIFT-based and the traditional point-based solvers.

\subsection{Essential Matrix Estimation}

For essential matrix estimation, we compare the 5PT algorithm (implemented in the Theia library~\cite{sweeney2015theia}) to the SIFT-based solver described in Section~\ref{sec:ess_solver}.
The solver used for fitting to a larger-than-minimal sample in GC-RANSAC is the 5PT algorithm. 
The inlier-outlier threshold is set to $0.75$ pixels and is normalized by the focal lengths. 

The cumulative distribution functions (CDF) of the rotation and translation errors (in degrees) and run-times (in seconds) of $\Ess$ estimation on the \num{69537} image pairs from the KITTI dataset are in Fig.~\ref{fig:ess_kitti_cdf}. 
The frame difference is denoted by color, \eg, image pairs ($I_i$, $I_{i+2}$) are considered for the green curve.  
The proposed solver yields almost exactly the same accuracy as the widely used point-based one while being \textit{significantly} faster as shown in the right plot.
For example, when the frame distance is $4$, GC-RANSAC with the point-based solver finishes earlier than $0.1$ seconds only on the ${\approx}17\%$ of the images pairs.
GC-RANSAC with the SIFT-based solver finishes faster than $0.1$ seconds in the $98\%$ of the cases. 
The results on the PhotoTourism dataset look similar in Fig~\ref{fig:ess_photo_cdf}. 
In this case, the proposed solver leads to comparable results to the 5PT algorithm and it is, again, significantly faster. 

The corresponding avg.\ errors, run-times and iteration numbers are reported in the first two rows of Table~\ref{table:results}. 
On KITTI, all methods have similar accuracy with the SIFT-based ones being \textit{five times} faster and \textit{real-time}.
On the PhotoTourism dataset, we show the median errors since it is significantly more challenging than KITTI and, thus, all methods fail on some pairs. 
Both the rotation and translation errors are similar for all solvers.
The run-time of the 3SIFT solver is \textit{eight times} lower than that of 5PT. 

\subsection{Fundamental Matrix Estimation}

For $\Fund$ estimation, we compare the 7PT algorithm~\cite{hartley2003multiple} to the SIFT-based solver described in Section~\ref{sec:ess_solver}.
The solver used for fitting to a larger-than-minimal sample in GC-RANSAC is the normalized 8PT algorithm. 
The inlier threshold is set to $0.75$ px. 

The CDFs of the rotation and translation errors (in degrees) and run-times (in seconds) of $\Fund$ estimation on the \num{69537} image pairs from KITTI are in Fig.~\ref{fig:fund_kitti_cdf}. 
Similarly as in the $\Ess$ estimation figure, the proposed solver yields almost exactly the same accuracy as the widely used point-based one while being \textit{significantly} faster as shown in the right plot.
The run-time difference is marginally smaller in this case due to the 7PT solver, used for $\Fund$ fitting, having fewer solutions than the 5PT algorithm. 
The results on the PhotoTourism dataset in Fig~\ref{fig:fund_photo_cdf} show that the proposed solver leads to the most accurate results while being three times faster than its point-based counterpart.

The corresponding average errors, run-times and iteration numbers are reported in the second two rows of Table~\ref{table:results}. 
On KITTI, all methods have similar accuracy while the SIFT-based solver is almost \textit{three times} faster than the point-based one.
On the PhotoTourism dataset, the SIFT-based solver leads to results superior to the point-based one both in terms of relative pose accuracy and run-time.

\subsection{Fundamental Matrix and Focal Length Estimation}

\begin{figure}[t!]
	\begin{center}
	\begin{subfigure}[t]{1.0\columnwidth}
	    \centering
		\includegraphics[width=0.30\columnwidth,trim={1mm 1mm 10mm 1mm},clip]{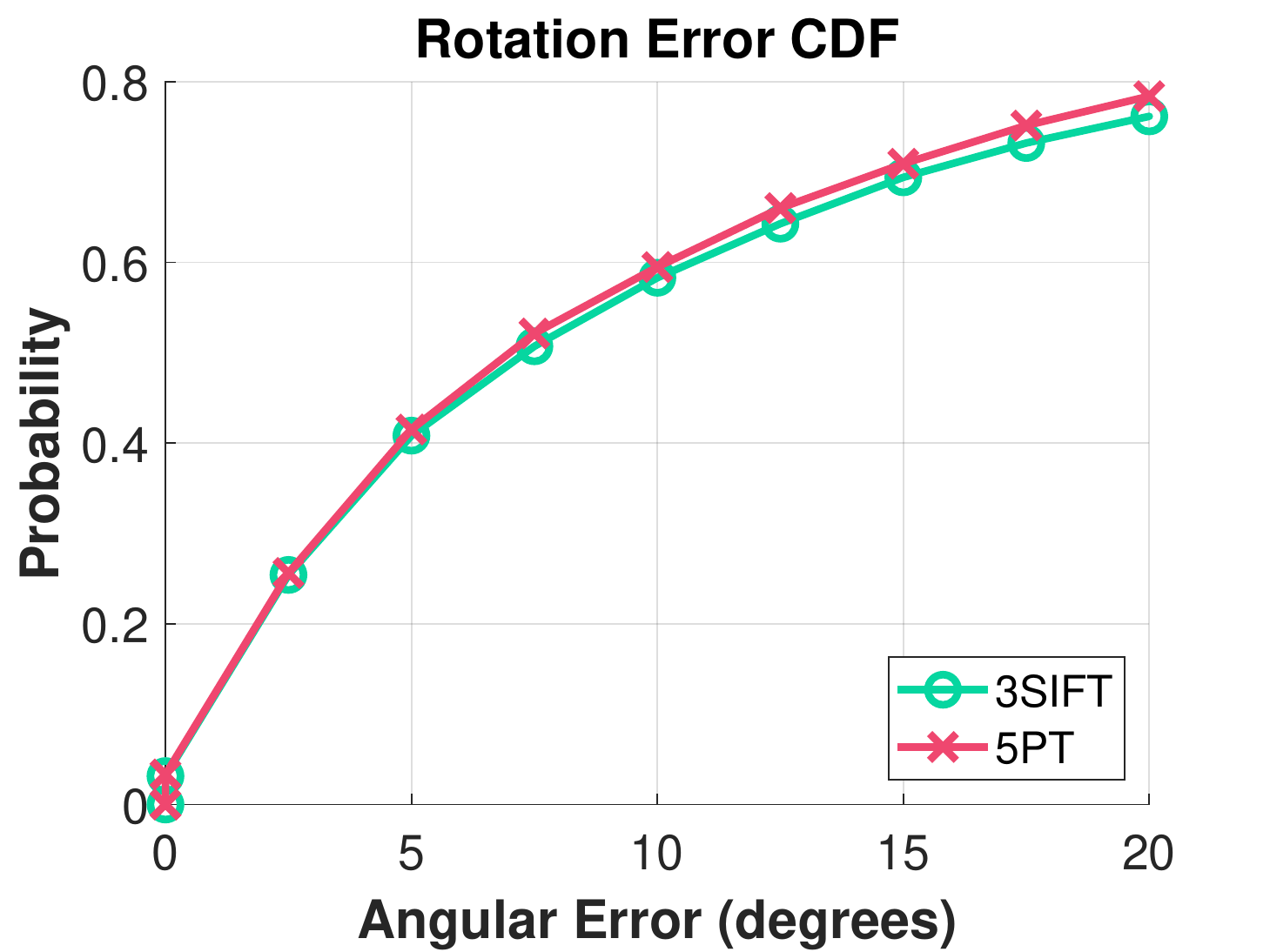}
		\includegraphics[width=0.30\columnwidth,trim={1mm 1mm 10mm 1mm},clip]{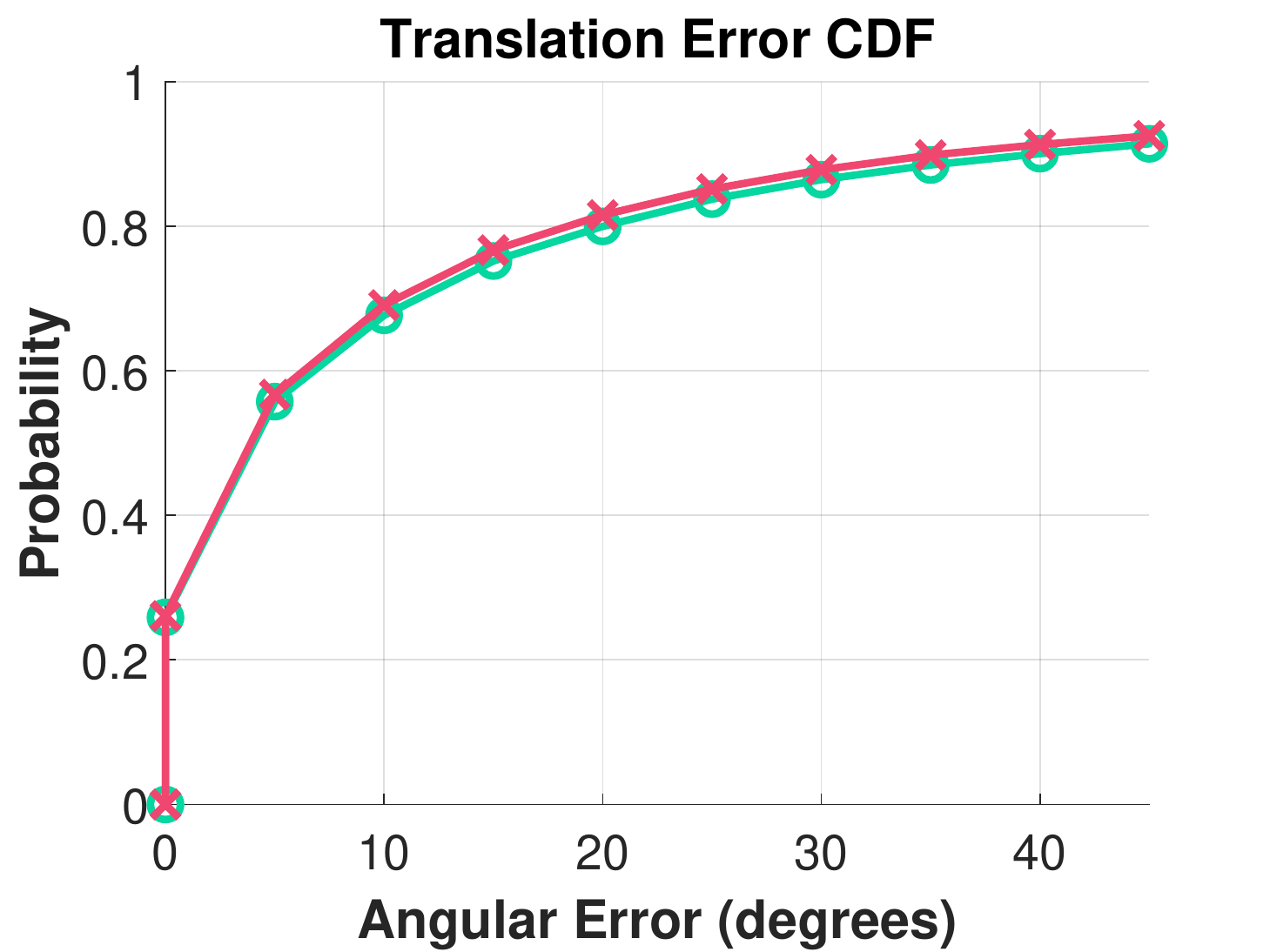}
		\includegraphics[width=0.30\columnwidth,trim={1mm 1mm 10mm 1mm},clip]{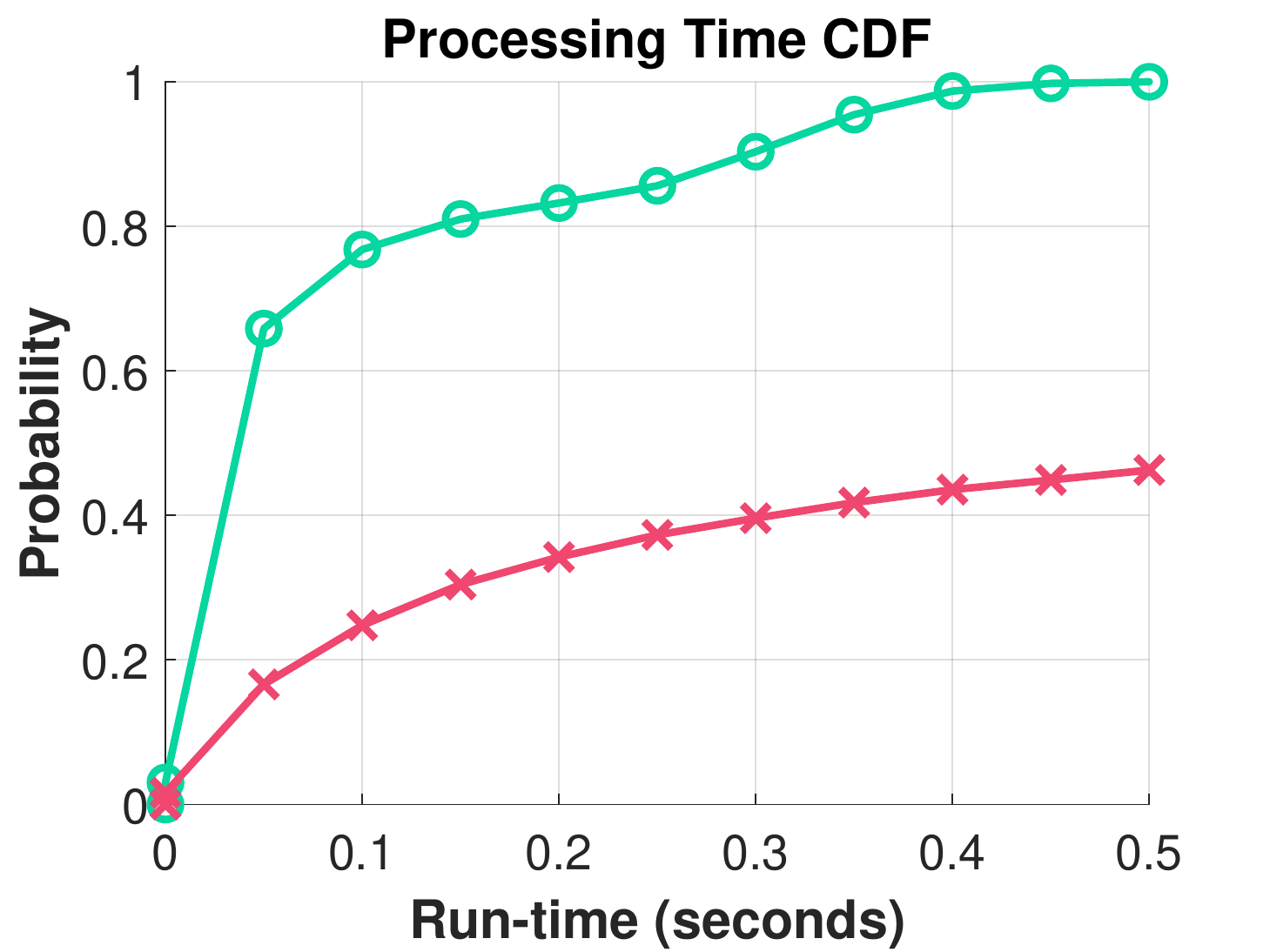}
		\caption{Essential matrix estimation}
		\label{fig:ess_photo_cdf}
	\end{subfigure}
	\begin{subfigure}[t]{1.0\columnwidth}
	    \centering
		\includegraphics[width=0.30\columnwidth,trim={1mm 1mm 10mm 1mm},clip]{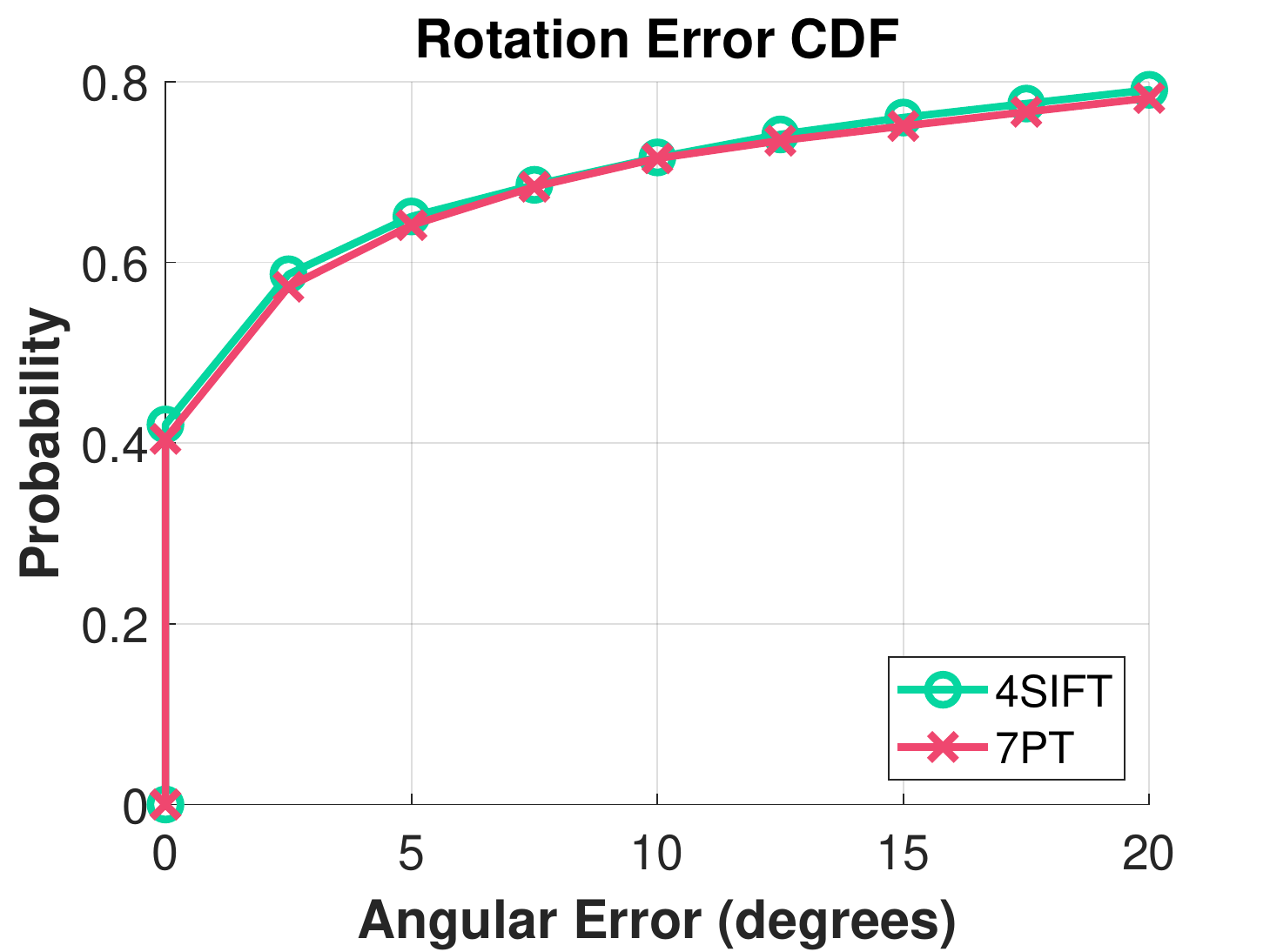}
		\includegraphics[width=0.30\columnwidth,trim={1mm 1mm 10mm 1mm},clip]{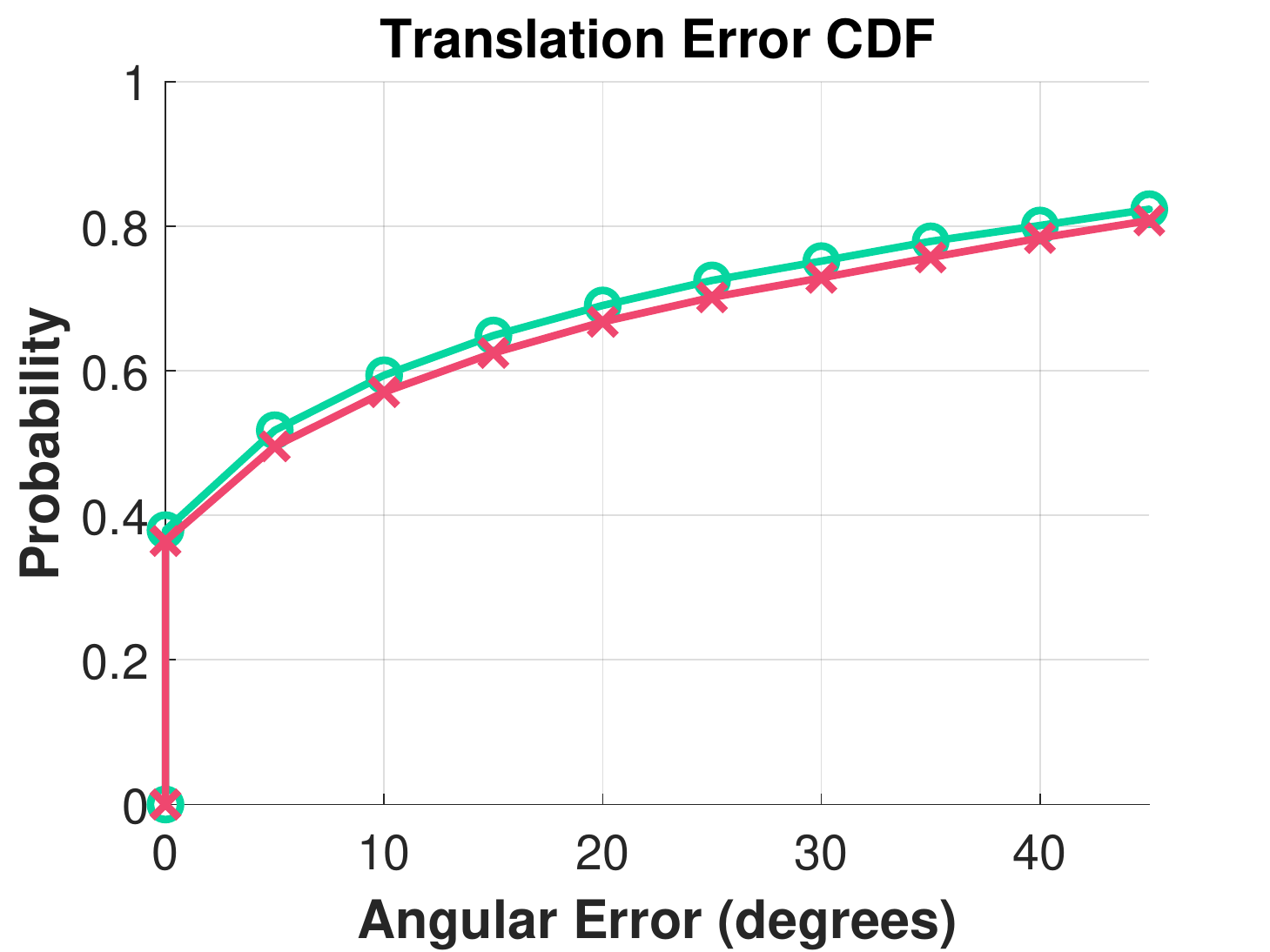}
		\includegraphics[width=0.30\columnwidth,trim={1mm 1mm 10mm 1mm},clip]{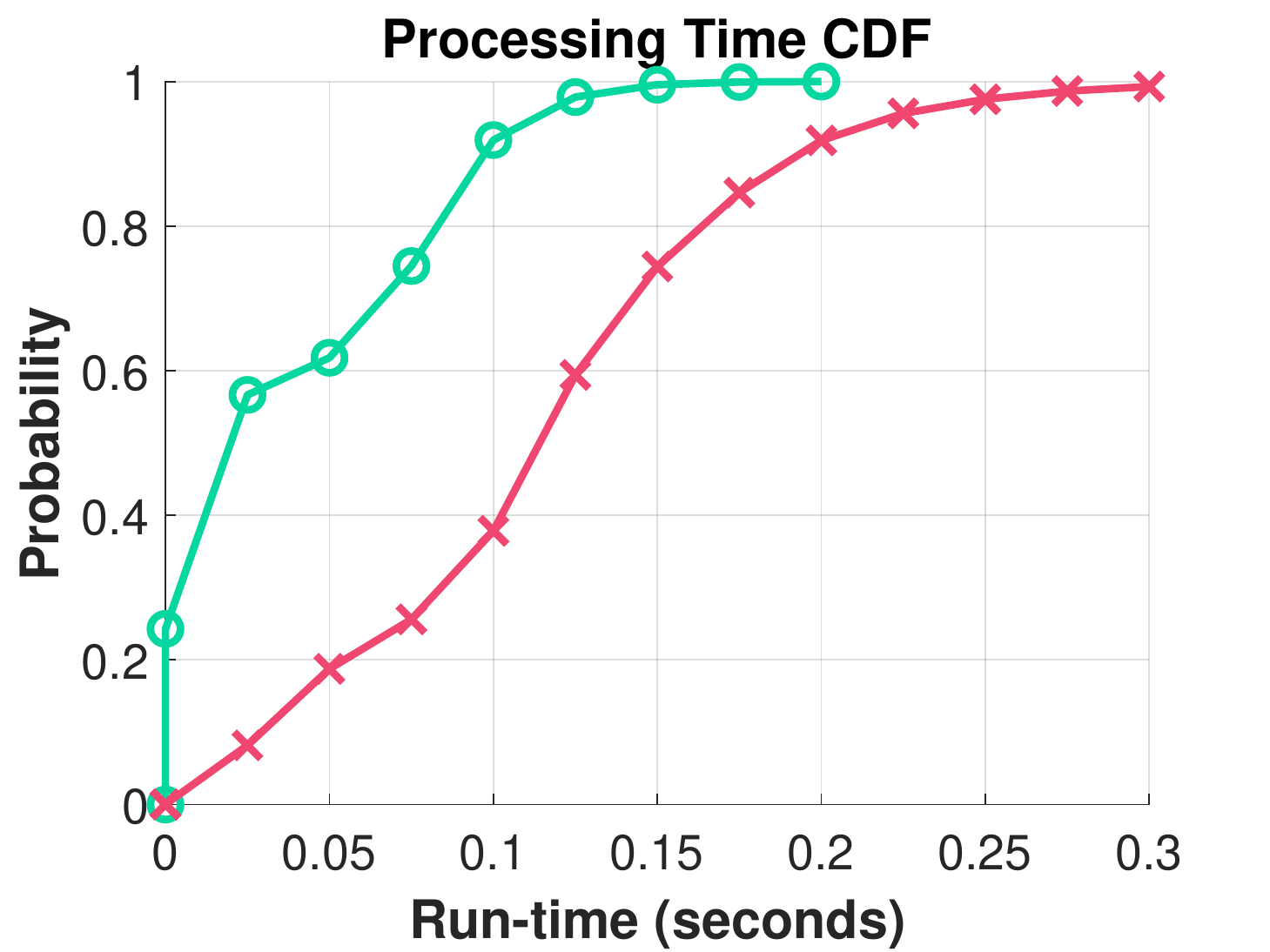}
		\caption{Fundamental matrix estimation}
		\label{fig:fund_photo_cdf}
	\end{subfigure}
	\begin{subfigure}[t]{1.0\columnwidth}
	    \centering
		\includegraphics[width=0.30\columnwidth,trim={1mm 1mm 10mm 1mm},clip]{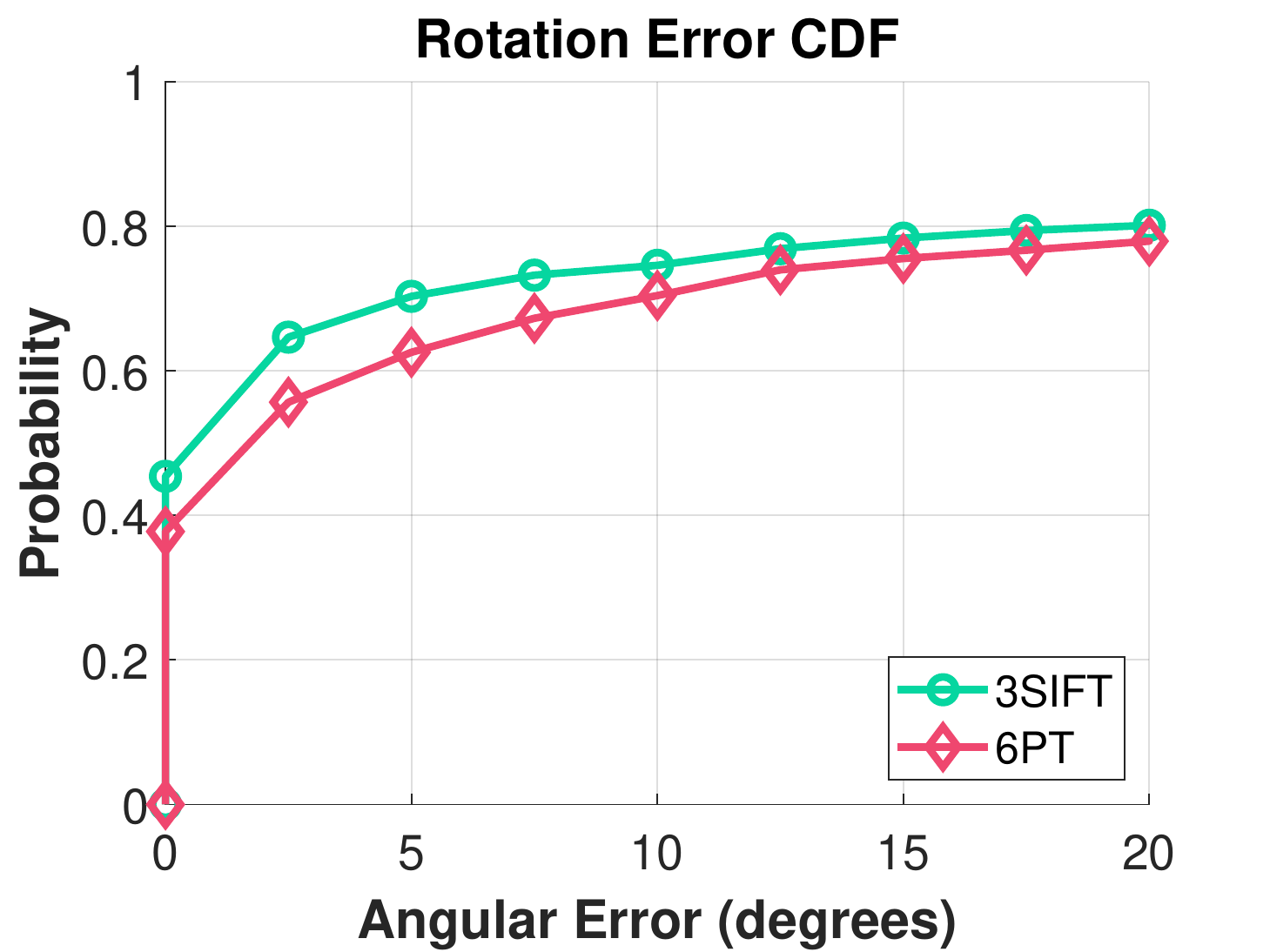}
		\includegraphics[width=0.30\columnwidth,trim={1mm 1mm 10mm 1mm},clip]{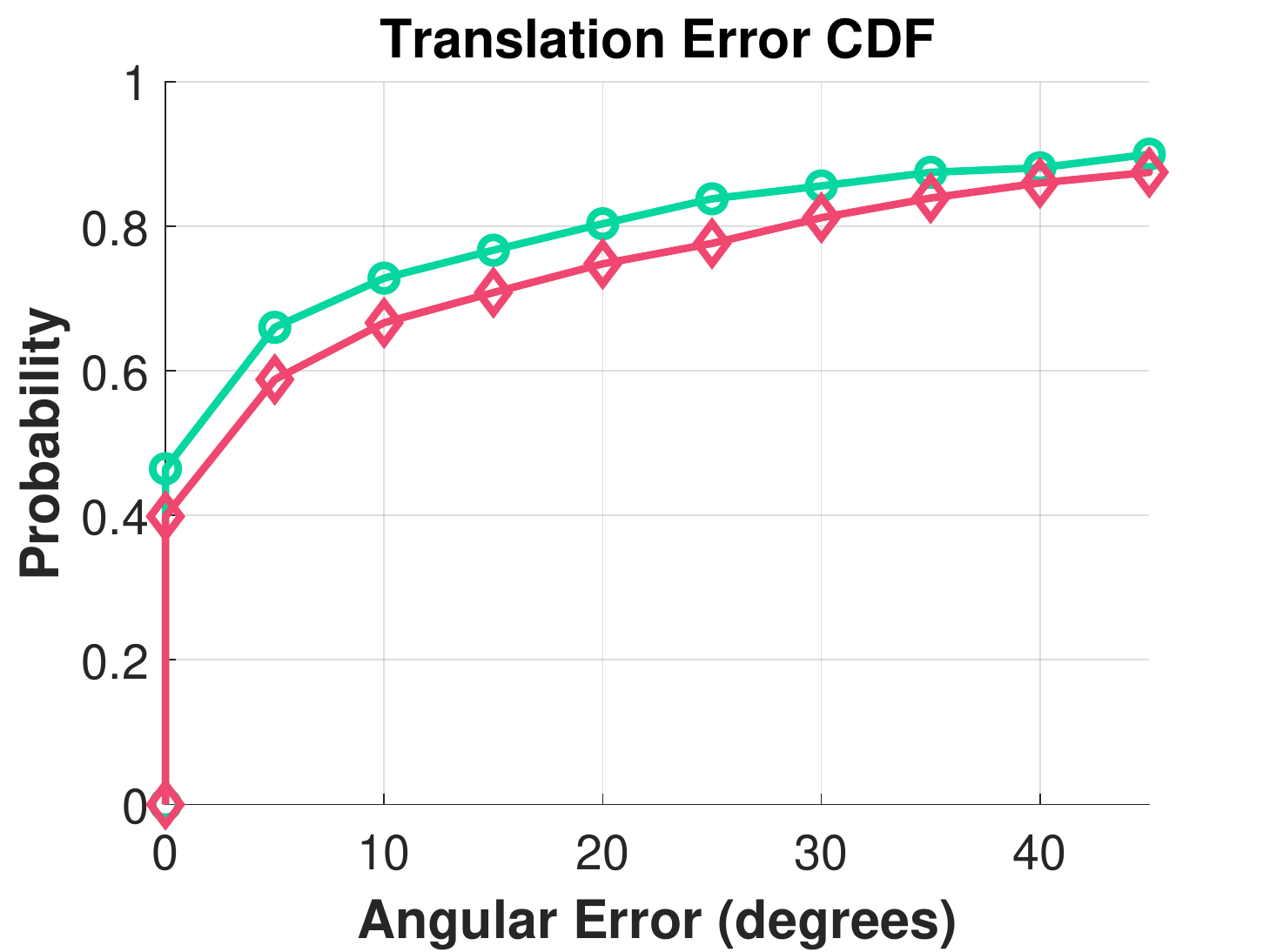}
		\includegraphics[width=0.30\columnwidth,trim={1mm 1mm 10mm 1mm},clip]{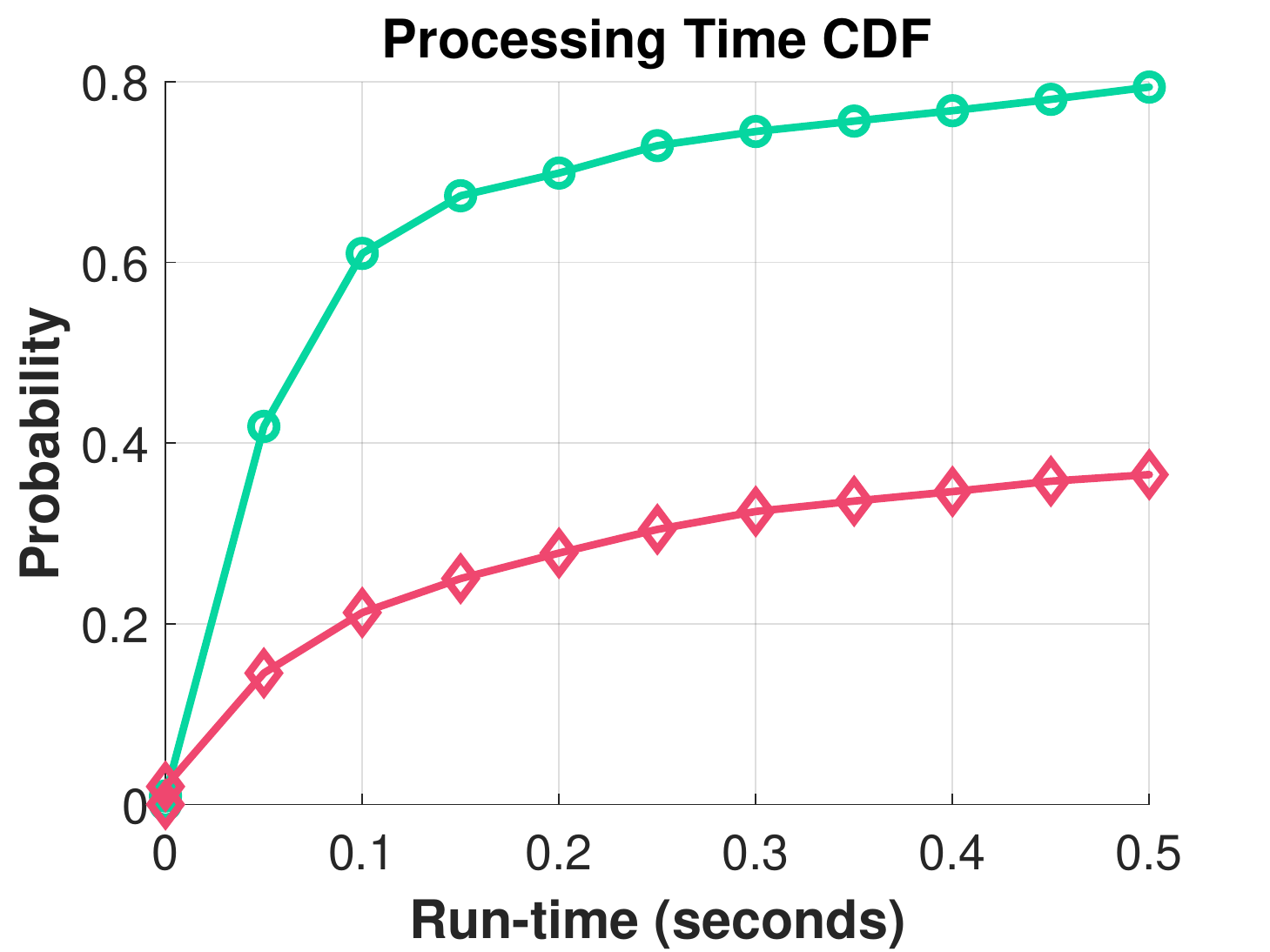}
		\caption{Fundamental matrix and focal length estimation}
		\label{fig:fFf_photo_cdf}
	\end{subfigure}
	\end{center}
	\caption{The cumulative distribution functions of the rotation and translation errors ($^\circ$) and run-times (secs) of epipolar geometry estimation by GC-RANSAC~\cite{barath2017graph} combined with point-based and the proposed SIFT-based minimal solvers on \num{9900} image pairs from the PhotoTourism dataset~\cite{cvpr2020ransactutorial}. }
\end{figure}

For $\Fund$ with focal length estimation, we compare the 6PT algorithm of~\cite{kukelovaCVPR2017} to the SIFT-based solver described in Section~\ref{sec:ess_solver}.
The inlier-outlier threshold is set to $0.75$ pixels. 
The CDFs of the rotation and translation errors (in degrees) and run-times (in seconds) on the \num{69537} image pairs from the KITTI dataset are in Fig.~\ref{fig:fFf_kitti_cdf}. 
Similarly as in the previous experiments, the proposed solver leads to almost exactly the same accuracy as the widely used point-based one while being \textit{significantly} faster as shown in the right plot.
%
The results on the PhotoTourism dataset in Fig~\ref{fig:fFf_photo_cdf} show that the proposed solver leads to increased accuracy compared to the 6PT solver while, also, being notably faster.
Note that, in order to use this solver, we used only those image pairs from the PhotoTourism dataset where the focal lengths are similar.

The corresponding average errors, run-times and iteration numbers are reported in the last two rows of Table~\ref{table:results}. 
The proposed solvers lead to the most accurate results while being the fastest by a large margin on both datasets.


\section{Conclusion}

We derive the general relationship of the epipolar geometry of perspective cameras and orientation and scale-covariant features. 
It is characterized by two linear equations, one from the point correspondence and one from the orientations and scales. 
These constraints can be used within \textit{all} existing relative pose solvers to halve the number of correspondences required for the estimation.
This leads to either similar or better accuracy while significantly accelerating the robust estimation -- by $4.3$ times, on average, on the tested popular computer vision problems.

\clearpage
%
%
\bibliographystyle{splncs04}
\bibliography{egbib}

\begin{thebibliography}{10}
\providecommand{\url}[1]{\texttt{#1}}
\providecommand{\urlprefix}{URL }
\providecommand{\doi}[1]{https://doi.org/#1}

\bibitem{barath2017phaf}
Barath, D.: {P-HAF}: Homography estimation using partial local affine frames.
  In: International Conference on Computer Vision Theory and Applications
  (2017)

\bibitem{barath2018five}
Barath, D.: Five-point fundamental matrix estimation for uncalibrated cameras.
  Conference on Computer Vision and Pattern Recognition  (2018)

\bibitem{cvpr2020ransactutorial}
Barath, D., Chin, T.J., Chum, O., Mishkin, D., Ranftl, R., Matas, J.: {RANSAC}
  in 2020 tutorial. In: CVPR (2020),
  \url{http://cmp.felk.cvut.cz/cvpr2020-ransac-tutorial/}

\bibitem{barath2017theory}
Barath, D., Hajder, L.: A theory of point-wise homography estimation. Pattern
  Recognition Letters  \textbf{94},  7--14 (2017)

\bibitem{barath2017graph}
Barath, D., Matas, J.: Graph-{C}ut {RANSAC}. In: Conference on Computer Vision
  and Pattern Recognition (2018)

\bibitem{barath2017focal}
Barath, D., Toth, T., Hajder, L.: A minimal solution for two-view focal-length
  estimation using two affine correspondences. In: Conference on Computer
  Vision and Pattern Recognition (2017)

\bibitem{barath2018approximate}
Barath, D.: Approximate epipolar geometry from six rotation invariant
  correspondences. In: International Conference on Computer Vision Theory and
  Applications (2018)

\bibitem{barath2018recovering}
Barath, D.: Recovering affine features from orientation-and scale-invariant
  ones. In: Asian Conference on Computer Vision (2018)

\bibitem{barath2018efficient}
Barath, D., Hajder, L.: Efficient recovery of essential matrix from two affine
  correspondences. IEEE Transactions on Image Processing  \textbf{27}(11),
  5328--5337 (2018)

\bibitem{barath2019homography}
Barath, D., Kukelova, Z.: Homography from two orientation-and scale-covariant
  features. In: Proceedings of the IEEE/CVF International Conference on
  Computer Vision. pp. 1091--1099 (2019)

\bibitem{barath2015optimal}
Barath, D., Molnar, J., Hajder, L.: Optimal surface normal from affine
  transformation. In: International Joint Conference on Computer Vision,
  Imaging and Computer Graphics Theory and Applications. SciTePress (2015)

\bibitem{barath2020making}
Barath, D., Polic, M., F{\"o}rstner, W., Sattler, T., Pajdla, T., Kukelova, Z.:
  Making affine correspondences work in camera geometry computation. In:
  European Conference on Computer Vision. pp. 723--740. Springer (2020)

\bibitem{bay2006surf}
Bay, H., Tuytelaars, T., Van~Gool, L.: {SURF}: Speeded up robust features.
  European Conference on Computer Vision  (2006)

\bibitem{Bentolila2014}
Bentolila, J., Francos, J.M.: Conic epipolar constraints from affine
  correspondences. Computer Vision and Image Understanding  (2014)

\bibitem{bentolila2014conic}
Bentolila, J., Francos, J.M.: Conic epipolar constraints from affine
  correspondences. Computer Vision and Image Understanding  (2014)

\bibitem{Geiger2012CVPR}
Geiger, A., Lenz, P., Urtasun, R.: Are we ready for autonomous driving? the
  {KITTI} vision benchmark suite. In: Conference on Computer Vision and Pattern
  Recognition (CVPR) (2012)

\bibitem{guan2021relative}
Guan, B., Zhao, J., Barath, D., Fraundorfer, F.: Relative pose estimation for
  multi-camera systems from affine correspondences. In: International
  Conference on Computer Vision. IEEE (2021)

\bibitem{hartley2012efficient}
Hartley, R., Li, H.: An efficient hidden variable approach to minimal-case
  camera motion estimation. Pattern Analysis and Machine Intelligence  (2012)

\bibitem{hartley2003multiple}
Hartley, R., Zisserman, A.: Multiple view geometry in computer vision.
  Cambridge University Press (2003)

\bibitem{koser2009geometric}
K{\"o}ser, K.: Geometric Estimation with Local Affine Frames and Free-form
  Surfaces. Shaker (2009)

\bibitem{kukelovaCVPR2017}
Kukelova, Z., Kileel, J., Sturmfels, B., Pajdla, T.: A clever elimination
  strategy for efficient minimal solvers. In: Conference on Computer Vision and
  Pattern Recognition (2017), http://arxiv.org/abs/1703.05289

\bibitem{li2006five}
Li, H., Hartley, R.: Five-point motion estimation made easy. In: International
  Conference on Pattern Recognition (2006)

\bibitem{li2006simple}
Li, H.: A simple solution to the six-point two-view focal-length problem. In:
  European Conference on Computer Vision. pp. 200--213. Springer (2006)

\bibitem{lowe1999object}
Lowe, D.G.: Object recognition from local scale-invariant features. In:
  International Conference on Computer vision (1999)

\bibitem{mikolajczyk2005comparison}
Mikolajczyk, K., Tuytelaars, T., Schmid, C., Zisserman, A., Matas, J.,
  Schaffalitzky, F., Kadir, T., Van~Gool, L.: A comparison of affine region
  detectors. International journal of computer vision  \textbf{65}(1-2),
  43--72 (2005)

\bibitem{mills2018four}
Mills, S.: Four-and seven-point relative camera pose from oriented features.
  In: International Conference on 3D Vision. pp. 218--227. IEEE (2018)

\bibitem{mishkin2015mods}
Mishkin, D., Matas, J., Perdoch, M.: {MODS}: Fast and robust method for
  two-view matching. Computer Vision and Image Understanding  (2015)

\bibitem{mishkin2018repeatability}
Mishkin, D., Radenovic, F., Matas, J.: Repeatability is not enough: Learning
  affine regions via discriminability. In: Proceedings of the European
  Conference on Computer Vision (ECCV). pp. 284--300 (2018)

\bibitem{Molnar2014}
Moln\'ar, J., Chetverikov, D.: Quadratic transformation for planar mapping of
  implicit surfaces. Journal of Mathematical Imaging and Vision  (2014)

\bibitem{morel2009asift}
Morel, J.M., Yu, G.: {ASIFT}: A new framework for fully affine invariant image
  comparison. SIAM journal on imaging sciences  \textbf{2}(2),  438--469 (2009)

\bibitem{nister2004efficient}
Nist{\'e}r, D.: An efficient solution to the five-point relative pose problem.
  Pattern Analysis and Machine Intelligence  (2004)

\bibitem{PerdochMC06}
Perdoch, M., Matas, J., Chum, O.: Epipolar geometry from two correspondences.
  In: International Conference on Pattern Recognition (2006)

\bibitem{Pritts2017RadiallyDistortedCT}
Pritts, J., Kukelova, Z., Larsson, V., Chum, O.: Radially-distorted conjugate
  translations. Conference on Computer Vision and Pattern Recognition  (2018)

\bibitem{pritts2018rectification}
Pritts, J., Kukelova, Z., Larsson, V., Chum, O.: Rectification from
  radially-distorted scales. In: Asian Conference on Computer Vision. pp.
  36--52. Springer (2018)

\bibitem{Raposo2016}
Raposo, C., Barreto, J.P.: Theory and practice of structure-from-motion using
  affine correspondences. In: Computer Vision and Pattern Recognition (2016)

\bibitem{raposo2016pi}
Raposo, C., Barreto, J.P.: $\pi$match: Monocular vslam and piecewise planar
  reconstruction using fast plane correspondences. In: European Conference on
  Computer Vision. pp. 380--395. Springer (2016)

\bibitem{stewenius2008minimal}
Stew{\'e}nius, H., Nist{\'e}r, D., Kahl, F., Schaffalitzky, F.: A minimal
  solution for relative pose with unknown focal length. Image and Vision
  Computing  \textbf{26}(7),  871--877 (2008)

\bibitem{sweeney2015theia}
Sweeney, C., Hollerer, T., Turk, M.: Theia: A fast and scalable
  structure-from-motion library. In: Proceedings of the 23rd ACM international
  conference on Multimedia. pp. 693--696 (2015)

\bibitem{IMC2020}
Trulls, E., Jun, Y., Yi, K., Mishkin, D., Matas, J., Fua, P.: Image matching
  challenge. In: CVPR (2020),
  \url{http://cmp.felk.cvut.cz/cvpr2020-ransac-tutorial/}

\bibitem{turkowski1990transformations}
Turkowski, K.: Transformations of surface normal vectors. In: Tech. Rep. 22,
  Apple Computer (1990)

\end{thebibliography}
\end{document}


\pagestyle{headings}
\mainmatter
\def\ECCVSubNumber{3998}  

\title{Supplementary Material:\\Relative Pose from SIFT Features} 

\titlerunning{ECCV-22 submission ID \ECCVSubNumber} 
\authorrunning{ECCV-22 submission ID \ECCVSubNumber} 
\author{Anonymous ECCV submission}
\institute{Paper ID \ECCVSubNumber}

\maketitle

\section{Constraints Relating Elements of $\matr{A}$ and SIFT Parameters}

In the main paper, in Section 3.1, we discussed constraints relating elements of the matrix $\matr{A}$ and SIFT parameters. The first set of constraints was derived from the decomposition of  $\matr{A}$ as the multiplication of the Jacobians of the projection functions w.r.t. the image directions in the two images. These equations have, after simplification, the following form (Eq. (11) in the main paper):
\begin{eqnarray}
    \begin{array}{lrclr}
    a_1 & = c_2 c_1 q_{u} - c_2 s_1 w + s_2 s_1 q_v, \\
    a_2 & = c_2 s_1 q_u + c_2 c_1 w - s_2 c_1 q_{v},\\
    a_3 & = s_2 c_1 q_{u} - s_2 s_1 w - c_2 s_1 q_v, \\
    a_4 & = s_2 s_1 q_u + s_2 c_1 w + c_2 c_1 q_{v},
    \end{array}
    \label{eq:DecomposedParameters}
\end{eqnarray}
where the unknowns are the affine parameters $a_1$, $a_2$, $a_3$, $a_4$, scales $q_u$, $q_v$ and shear $w$. Angles $\alpha_1$ and $\alpha_2$ are known from the SIFT features and $c_i = \cos(\alpha_i)$ and $s_i = \sin(\alpha_i)$.

In addition to these constraints, there are two more constraints relating SIFT parameters and $\matr{A}$. 
First, the uniform scales of the SIFT features are proportional to the area of the underlying image region and, therefore, the scale change provides constraint (Eq. (12) in the main paper):
\begin{equation}
    \small
    \det \textbf{A} = \det \left(\textbf{R}_2 \mat U \textbf{R}_1^{\trans} \right) = \det \mat U = q_u q_v = \frac{q_2^2}{q_1^2},
    \label{eq:ScaleDependency}
\end{equation}
where $q_1$ and $q_2$ are the SIFT scales in the two images. 
Second, the oriented circles centered on the point correspondence provide an additional constraint of the form (Eq. (13) in the main paper):
\begin{equation}
    \small
    q_1 \textbf{A} 
    \begin{bmatrix}
        \cos(\alpha_1) \\
        \sin(\alpha_1)
    \end{bmatrix} = q_2 \begin{bmatrix}
        \cos(\alpha_2) \\
        \sin(\alpha_2)
    \end{bmatrix}.
    \label{eq:CircleConstraint}
\end{equation}
The constraints~\eqref{eq:ScaleDependency} and~\eqref{eq:CircleConstraint} can be rewritten as
   \begin{eqnarray}
      \small
    \label{eq:constraint1}
    a_2 a_3 - a_1 a_4 + q^2 &=& 0,\\
    \label{eq:constraint2}
    a_3 c_1+a_4 s_1 - s_2 q &=& 0, \\
    \label{eq:constraint3}
    a_1 c_1+a_2 s_1 - c_2 q &=& 0,
\end{eqnarray}
where   $q= \frac{q_2}{q_1}$.

 Here, we show  that constraints~\eqref{eq:constraint1}-\eqref{eq:constraint3} constitute all constraints that relate the elements of $\matr{A}$ and the measured orientations $\alpha_i$ and scales $q_i,\,i=1, 2$ of the features in the first and second images. In other words, equations~\eqref{eq:DecomposedParameters} are not adding any additional constraints relating the elements of $\matr{A}$ and the measured orientations $\alpha_i$ and scales $q_i,\,i=1, 2$ to constraints~\eqref{eq:constraint1}-\eqref{eq:constraint3}.


To prove this, we first define the ideal $I$~\cite{cox2005} generated by polynomials~\eqref{eq:DecomposedParameters},\eqref{eq:ScaleDependency},\eqref{eq:CircleConstraint} %
and trigonometric identities $c_i^2 + s_i^2 = 1$ for $i \in \{1,2\}$. 
To transform~\eqref{eq:ScaleDependency} to a polynomial equation, we substitute $q= \frac{q_2}{q_1}$ and add a constraint $q_1 q - q_2 =0$ to the ideal. 
Moreover, we ensure $q_1 \neq 0$ by saturating the ideal with $q_1$\footnote{Note that geometrically $q_1 \neq 0$ and $q_2 \neq 0$, but algebraically it is sufficient to remove solutions $q_1 = 0$ by saturating the ideal with $q_1$. Saturating the ideal with both $q_1$ and $q_2$ will not affect the solutions.}.
Note that here we consider all elements of these polynomials, including $q$, $q_i$, $c_i$ and $s_i$, as unknowns. 
Then we compute the generators of the elimination ideal $I_1 = I \cap \mathbb{C}[a_1,a_2,a_3,a_4,q,s_1,c_1,s_2,c_2]$~\cite{cox2005}.
The generators of $I_1$ do not contain $q_u$, $q_v$ and $w$ and, instead, of $q_1$ and $q_2$ they directly contain $q= \frac{q_2}{q_1}$. These generators are exactly  equations~\eqref{eq:constraint1}-\eqref{eq:constraint3} together with trigonometric identities $c_i^2 + s_i^2 = 1$ for $i \in \{1,2\}$. This means that constraints~\eqref{eq:DecomposedParameters} do not add any additional information and the elimination ideal is directly generated by~\eqref{eq:ScaleDependency} and~\eqref{eq:CircleConstraint}, or equivalently by~\eqref{eq:constraint1}-\eqref{eq:constraint3}\footnote{Note that for correct derivations all steps including saturation and adding trigonometric identities to the ideal are important.}.
Generators~\eqref{eq:constraint1}-\eqref{eq:constraint3} can be computed using a computer algebra system, \eg,\ {\tt Macaulay2}~\cite{M2}. The input code for {\tt Macaulay2} is included in the supplementary material as a separate constraints$\_$A$\_$SIFT.m2 file.

On the other hand, the constraints~\eqref{eq:DecomposedParameters} and~\eqref{eq:ScaleDependency} that were used for derivation in~\cite{barath2019homography}, are not covering all constraints that relate the elements of $\matr{A}$ and the measured orientations $\alpha_i$ and scales $q_i,\,i=1, 2$. This can be easily proved. By eliminating $q_u$, $q_v$ and $w$ from the ideal generated by~\eqref{eq:DecomposedParameters},~\eqref{eq:ScaleDependency} and trigonometric identities $c_i^2 + s_i^2 = 1$ for $i \in \{1,2\}$, using the elimination ideal technique~\cite{kukelovaCVPR2017}\footnote{Note, that here we again substitute $q= \frac{q_2}{q_1}$, add the constraint $q_1 q - q_2 =0$ to the ideal and saturate the ideal with $q_1$. }, we obtain two generators of the elimination ideal. One generator is directly~\eqref{eq:constraint1}, \ie constraint~\eqref{eq:ScaleDependency}, and the second one has the form 
 \begin{eqnarray}
    \small
    \label{eq:constraint_old}
    c_1 s_2 a_1 + s_1 s_2 a_2 - c_1 c_2 a_3 - c_2 s_1 a_4 = 0.
\end{eqnarray}
The constraints~\eqref{eq:constraint1} and~\eqref{eq:constraint_old} are constraints proposed in~\cite{barath2019homography}

The constraint 
\eqref{eq:constraint_old} is a linear combination of constraints~\eqref{eq:constraint2} and~\eqref{eq:constraint3} with coefficients $-c_2$ and $s_2$. This means that if~\eqref{eq:constraint2} and~\eqref{eq:constraint3}  vanish also~\eqref{eq:constraint_old} vanishes. However, the opposite is not true. If~\eqref{eq:constraint_old} vanishes the constraints~\eqref{eq:constraint2} and~\eqref{eq:constraint3} do not need to vanish.  This means that constraints derived in~\cite{barath2019homography}, \ie, constraints~\eqref{eq:constraint1} and~\eqref{eq:constraint_old},  do not cover all constraints that relate the elements of $\matr{A}$ and the measured orientations $\alpha_i$ and scales $q_i,\,i=1, 2$. For deriving all constraints, the constraint~\eqref{eq:CircleConstraint} is important.

\section{SIFT Epipolar Constraint}
In the main paper, in Section 3.2., we derived a new constraint relating epipolar geometry and the measured orientations $\alpha_i$ and scales $q_i,\,i=1, 2$ of covariant features in the first and second images. For this purpose, we used elimination ideal technique~\cite{kukelovaCVPR2017}. 
The input code for {\tt Macaulay2} used to compute this constraint, \ie, the generator of the elimination ideal $J_1$ is provided as a separate {\tt Macaulay2} constraint$\_$F$\_$SIFT.m2 file.

%

%
%
\bibliographystyle{splncs04}
\bibliography{egbib}